\documentclass[preprints,article,accept,pdftex,moreauthors]{Definitions/mdpi}
\firstpage{1} 
\pubvolume{1}
\issuenum{1}
\articlenumber{0}
\pubyear{2022}
\copyrightyear{2022}
\datereceived{} 
\dateaccepted{} 
\datepublished{} 
\hreflink{https://doi.org/} 

\usepackage{makecell}
\newcommand{\tabincell}[2]{\begin{tabular}{@{}#1@{}}#2\end{tabular}}
\usepackage{multirow}
\usepackage{indentfirst}
\usepackage{pifont}
\usepackage{amssymb}
\usepackage{amsmath}
\usepackage{amsfonts}
\usepackage{graphicx}
\usepackage[misc]{ifsym}
\usepackage[final]{changes}
\definechangesauthor[name={Authors}, color=red]{A.} 
\usepackage{nomencl}
\makenomenclature

\Title{Co-visual pattern augmented generative transformer learning for automobile geo-localization}

\TitleCitation{Co-visual pattern augmented generative transformer learning for automobile geo-localization}

\Author{Jianwei Zhao$^{1,2}$, Qiang Zhai$^{1,2,*}$\orcidB{}, Pengbo Zhao$^{3}$, Rui Huang$^{1,2}$ and Hong Cheng$^{1,2}$\orcidC{}}

\AuthorNames{Jianwei Zhao, Qiang Zhai, Pengbo Zhao, Rui Huang, and Hong Cheng}

\AuthorCitation{Jianwei, Z.; Qiang, Z.; Pengbo Z.; Rui H.; Hong, C.}

\address{%
$^{1}$ \quad School of Automation Engineering, University of Electronic Science and Technology of China, CN \\
$^{2}$ \quad Center for Robotics, University of Electronic Science and Technology of China, CN \\
$^{3}$ \quad Northwestern University, USA
}

\corres{Correspondence: qzhai@std.uestc.edu.cn}

\abstract{
Geolocation is a fundamental component of route planning and navigation for unmanned vehicles, but GNSS-based geolocation fails under denial-of-service conditions. Cross-view geo-localization (CVGL), which aims to estimate the geographical location of the ground-level camera by matching against enormous geo-tagged aerial (\emph{e.g.}, satellite) images, has received lots of attention but remains extremely challenging due to the drastic appearance differences across aerial-ground views. In existing methods, global representations of different views are extracted primarily using Siamese-like architectures, but their interactive benefits are seldom taken into account. In this paper, we present a novel approach using cross-view knowledge generative techniques in combination with transformers, namely mutual generative transformer learning (MGTL), for CVGL. Specifically, by taking the initial representations produced by the backbone network, MGTL develops two separate generative sub-modules---one for aerial-aware knowledge generation from ground-view semantics and vice versa---and fully exploits the entirely mutual benefits through the attention mechanism. Moreover, to better capture the co-visual relationships between aerial and ground views, we introduce a cascaded attention masking algorithm to further boost accuracy. Extensive experiments on challenging public benchmarks, \emph{i.e.}, {CVACT} and {CVUSA}, demonstrate the effectiveness of the proposed method which sets new records compared with the existing state-of-the-art models. Our code will be available upon acceptance.
}


\begin{document}

\section{Introduction}\label{sec:1}
\begin{figure}[t!]
  \begin{center}
      \includegraphics[width=0.8\linewidth]{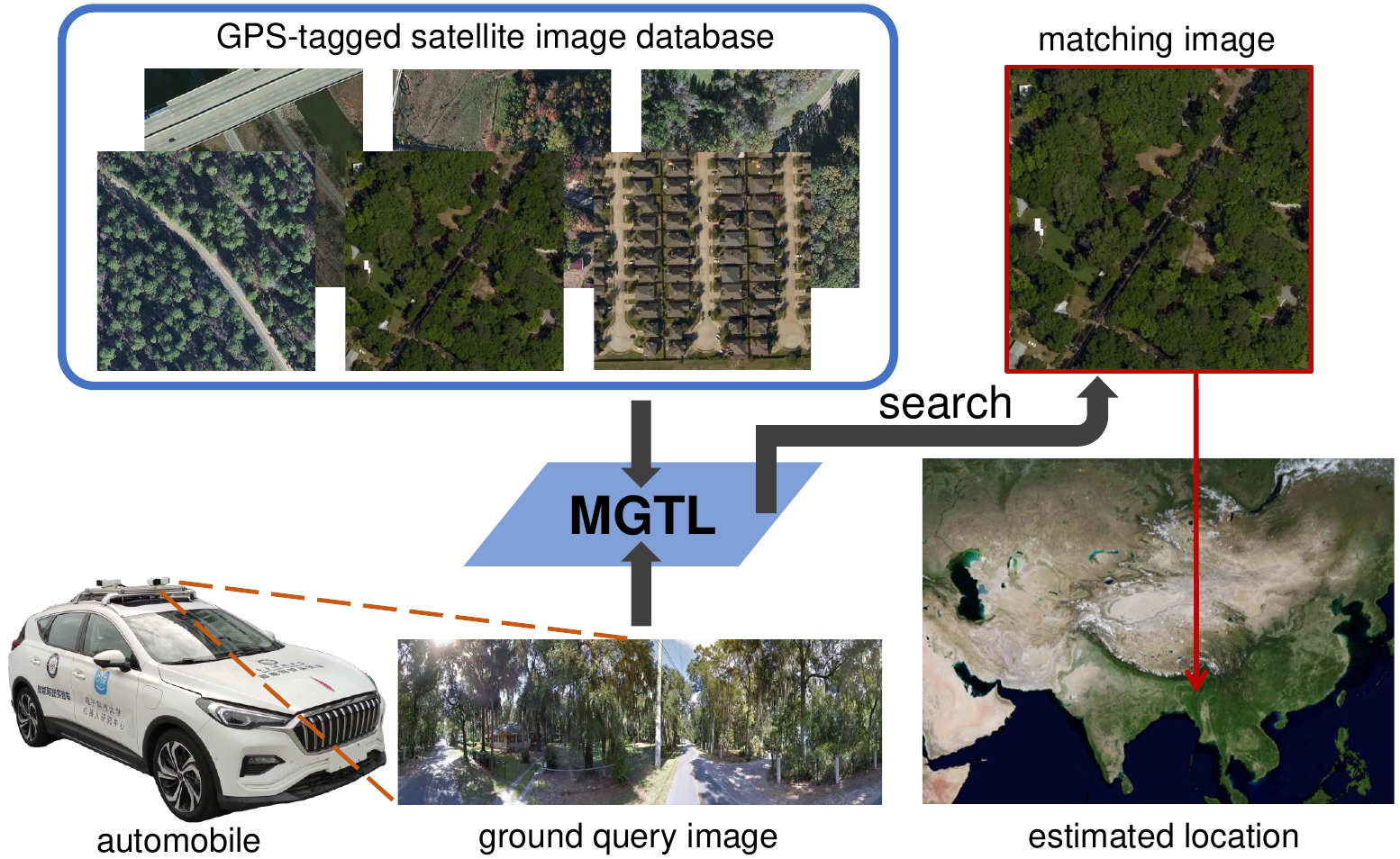}
  \end{center}
  \caption{The schematic diagram of image matching-based cross-view geo-localization.}
\label{fig:1}
\end{figure}

{Geolocation identification of automobiles has been a topic of growing interest in recent years due to its potential applications in navigation and route planning for intelligent vehicles~\cite{saurer2016image, 2012Satellite, xiao2020multimodal, wang2022satellite, thoma2019mapping,roy2020uav, hu2020image}. Conventionally, obtaining the geographical location of a vehicle through Global Navigation Satellite Systems (GNSS) has been a convenient and cost-effective method. However, GNSS signals are prone to being unreliable or unavailable due to the presence of dense high-rise obstacles, network failures, \emph{etc}. For example, such scenarios as dense primordial forests and crowded buildings are shown in Figure~\ref{fig:1}. Fortunately, current satellite imagery can cover most outdoor scenarios where automobiles involve and are easily collected offline in advance through open services like Google Maps. To overcome this limitation, the use of registered ground-satellite image retrieval for geographical location estimation has gained increasing attention~\cite{arandjelovic2016netvlad, workman2015location, vo2016localizing, hu2018cvm, regmi2019bridging, zhu2022transgeo, yang2021cross}. This method involves the comparison of the visual data obtained from the vehicle with geo-tagged references stored in a database, resulting in the estimation of the geographical location which is aligned with the closest reference, the pipeline is schematically illustrated in Figure~\ref{fig:1}.}

Typically, geolocation involves the collection of perspectives from sites previously visited by vehicles. Upon subsequent revisits, these views can be compared with similar scene content, constituting a loop closure detection process. In the event of satellite signal failure, the agent is required to determine its position by analyzing the contextual scene. Such methodologies are designed to mitigate ambiguity by exploring and encoding contextual information and deep semantics. \added[id=A., comment={Writing polishing.}]{The images are encoded by identical or Siamese-like backbone networks, followed by the nearest neighbor matching. Thus, the geolocation task is akin to image retrieval, albeit with a primary focus on capturing and leveraging geometric and structural information of the environmental features that constitute the scene. Such information may include but is not limited to, edge and corner features, shapes, and their relative positions, all of which are fundamental to effective geolocation}. Consequently, geolocation requires a more nuanced understanding of the scene content than traditional image retrieval, as it must incorporate this rich geometric and structural information into its matching algorithm to achieve accurate results. Overfeat~\cite{chen2014convolutional} was a pioneering deep learning-based study in the field and inspired a series of improvements~\cite{arandjelovic2016netvlad, Xin2019LLRN, Khaliq2022MultiRes, yu2019spatial, latif2018addressing}. {To construct a reference data set with GPS information, these approaches examine the ground-to-ground matching procedure for localization by gathering views at diverse locations at different times, seasons, and weather conditions, as exemplified by Google Street View, a widely used application. During the localization phase, views with unknown locations are matched with reference sets to estimate their locations. Despite their effectiveness, these methods are labor-intensive and cannot locate places that are not in the reference dataset. Therefore, researchers are striving to establish the interconnectivity between satellite views and ground views by extracting the intrinsic similarities between the two view types, namely cross-view geo-localization (CVGL), which increases the generalization performance of the location model.} Owing to the dissimilar imaging perspective between satellite and ground views, the appearance of content varies significantly, posing a substantial challenge in achieving cross-view localization. Nonetheless, researchers have made remarkable strides in devising Siamese-like networks that contain two distinct branches responsible for encoding each view independently~\cite{2015Semantic, 2016Semantic, yang2021cross, zhu2021vigor, shi2019spatial, shi2020optimal, wang2022transgcnn, 2021Each, wang2022learning, 
 zhu2023simple, zhang2022cross}. While the relationship between different views provides a significant impetus for cross-view localization, several challenges persist. \textbf{First}, semantic consistency between views is not fully leveraged. Current methods typically utilize Siamese-like networks for independent encoding of cross-view views but often neglect the high-order consistency semantics of view content, which is essential for matching ground and satellite images. \textbf{Second}, co-visual relationships between views are not explicitly accounted for. The perspective disparities between ground and satellite views limit co-visual relationships exploring, with the latter typically encompassing a more extensive scope, thus using the whole image for coding would yield suboptimal accuracy. \textbf{Third}, deep contextual semantic mining is not yet sufficient. As the interaction between views remains unconsidered, the existing methods fail to fully explore contextual semantics.

\begin{figure}[ht!]
  \begin{center}
      \includegraphics[width=0.75\linewidth]{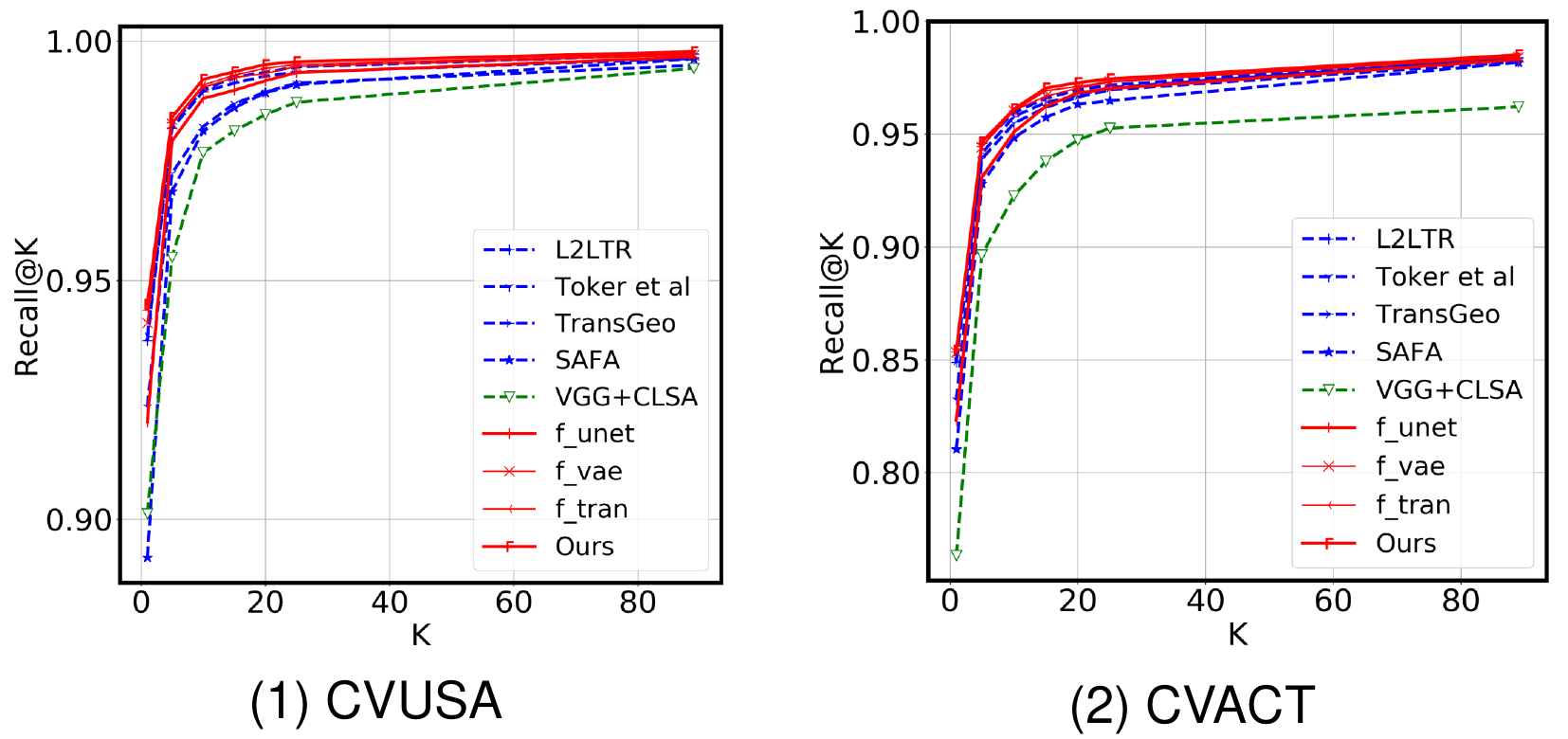}
  \end{center}
  \caption{The proposed MGTL outperforms existing approaches.}
\label{fig:2}
\end{figure}



To address the above deficiencies, we present a novel mutual generative transformer learning (MGTL) for the CVGL task. We first revisit the attention learning strategy and propose a novel {cascaded attention masking algorithm} to make the network reasoning on the co-visual patterns between ground and satellite views. Then, two symmetrical generative sub-modules, \emph{i.e.,} {Ground-to-Satellite (G2S) and Satellite-to-Ground (S2G) are thoughtfully designed to generate the simulated cross-view knowledge and to capitalize on the mutual benefits across views. Specifically, S2G takes the aerial semantics and skillfully simulates the ground-aware knowledge, and vice versa. \added[id=A., comment={Writing polishing.}]{Subsequently, the view-specific simulated knowledge is applied to strengthen the current view features via attention learning and all the sub-components work in concert within a transformer-based framework to accomplish the CVGL task.} The experiment results on several public challenging benchmarks unequivocally establish the superiority of our proposal and the contributions of the proposed MGTL can be summarized as follows:}

\begin{itemize}
\item {\bf A novel cross-view knowledge guided learning approach for CVGL.} To the best of our knowledge, the MGTL is the first attempt to build the mutual interaction between ground-level and aerial-level patterns in the CVGL community. Unlike existing transformer-based CVGL models that only perform self-attentive reasoning in the respective view, our proposed MGTL performs cross-knowledge information to achieve more representative high-order features.
\item{\bf Cascaded attention-guided masking to exploit the co-visual patterns.} Instead of treating patterns in aerial and ground views equally, we developed an attention-guided exploration algorithm to make the network reasoning based on the co-visual patterns, which further improves performance.

\item {\bf State-of-the-art localization accuracy on widely-used benchmarks.} The proposed MGTL outperforms existing deep models on various datasets, \emph{i.e.,} \emph{CVUSA}~\cite{workman2015wide} and \emph{CVACT}~\cite{liu2019lending}.

\end{itemize}

\section{RELATED WORKS}\label{sec:2}

{\bf Visual Place Recognition:} Visual place recognition (VPR) involves matching ground-to-ground images and is a crucial aspect of vision-based navigation and localization, particularly in the field of autonomous driving. The problem tackled by VPR is to determine the current camera's location in the existing image database and the task becomes challenging due to factors such as seasonal variations and dynamic visual object changes. A primary approach to overcoming this challenge involves extracting high-level feature descriptors from input images and comparing them based on distance. To improve VPR performance, Arandjelovic \emph{et al.}~\cite{arandjelovic2016netvlad} modified the traditional Non-differentiable operation in Vector of Locally Aggregated Descriptors (VLAD) and incorporated it into CNN-based networks to develop an end-to-end trainable VLAD descriptor, named NetVLAD. Following the success of NetVLAD, several variants~\cite{yu2019spatial, Khaliq2022MultiRes} have been proposed. {To exploit the multi-scale information, Spatial pyramid-enhanced NetVLAD (SPE-NetVLAD)~\cite{yu2019spatial} integrated multi-scale features in the training phase by cascading encoding features with varying scales in the final convolutional layer of NetVLAD to improve the performance of VPR. Multi-Resolution NetVLAD (MultiRes-NetVLAD)~\cite{Khaliq2022MultiRes} utilized low-resolution image pyramid coding and presented a Multi-Resolution Residual Aggregation scheme to enhance the NetVLAD learning feature representation capability.} In addition, to address the issue of seasonal and time-of-day variations, Latif \emph{et al.}~\cite{latif2018addressing} approached the VPR problem as a region translation task. A pair of coupled Generative Adversarial Networks (GANs) was utilized to generate the appearance of one domain from another without requiring image-to-image correspondences across the domains. These classical solutions in the field of VPR offer valuable insights for addressing the challenge of cross-view image matching.

{\bf Cross-View Geo-localization:} 
{Current cross-view geo-localization (CVGL) pipelines utilize a Siamese-like neural network to extract feature representations from each view, followed by the definition of a metric that places the embedding features of cross-view images in close proximity based on their GPS coordinates.} The primary obstacle in CVGL tasks is the significant appearance gap between ground and aerial views caused by changes in viewpoint~\cite{zhu2021geographic}. Satellite-view images are typically composed of satellite images captured by specialized panchromatic and multispectral cameras on board satellites, whereas ground-view images consist of panoramic images taken using handheld or vehicular optical cameras. These two images have different imaging principles and shooting angles, leading to stark differences in image appearance, such as the representation of visual objects and their spatial layout. This problem is further exacerbated by the large time intervals between image acquisition. Prior work has mainly addressed this issue by focusing on extracting viewpoint-invariant features~\cite{zhu2022r2fd2, zhu2023simple, zhang2022cross} or applying viewpoint transformation~\cite{regmi2018cross, lu2020geometry, ding2020cross}. The former involves designing effective network architectures that can extract invariant features across views. Workman \emph{et al.}~\cite{workman2015location} proposed a convolutional neural network (CNN) to learn a joint semantic feature representation for aerial and ground-level imagery, while Lin \emph{et al.}~\cite{Lin2015Learning} introduced a Siamese-like network followed by Euclidean distance calculation to measure cross-view feature representation similarity. More recently, Hu \emph{et al.}~\cite{hu2018cvm} utilized NetVLAD to encode global descriptors and a Siamese-like CNN-based network to extract local feature descriptors for more robust representation learning. Sun {et al.}~\cite{sun2019GeoCapsNet} further presented a pure convolutional network equipped with capsule layers to model the spatial feature hierarchies. In contrast, to address the imagery geometric gap caused by viewpoint differences, Shi \emph{et al.}~\cite{shi2019spatial, 2019Ground} used polar transform and attention mechanisms to pre-process satellite imagery, which has been shown to be highly effective. Recently, Yang \emph{et al.}~\cite{yang2021cross} and Zhu \emph{et al.}~\cite{zhu2022transgeo} proposed transformer-based methods, leveraging self-attention mechanisms to model global dependencies. Zhu \emph{et al.}~\cite{zhu2022transgeo} introduced a novel attention-based masking mechanism to remove redundant areas in satellite images, reducing interference in matching performance. The latter approach involves exploring ways to synthesize realistic cross-domain imagery using viewpoint transformation. {Ren \emph{et al.}~\cite{ren2021cascaded} proposed a cascaded cross mlp-mixer GAN (CrossMLP) module to extract latent mapping cues between cross-view imagery, while Toker \emph{et al.}~\cite{toker2021coming} developed a GAN-based multi-task architecture to synthesize realistic street views from satellite images.} {However, existing methods lack mutual learning across views and fail to consider the inter-dependencies between latent features in different network branches. In this paper, we propose a novel approach that integrates cross-view knowledge generative tactics into the transformer architecture, referred to as mutual generative transformer learning. This approach leverages mutual learning across different views to improve the feature representation capability and retrieval performance.}

{\bf Vision Transformer:} The Transformer~\cite{A2017Attention} has gained widespread use in the field of natural language processing (NLP) due to its excellent global modeling capability and self-attention mechanism, as demonstrated by its superior properties~\cite{A2017Attention}. The self-attention mechanism is based on the calculation of dot-product similarity by query and key, which are then multiplied with value, where query, key, and value represent different embedding spaces computed by the input feature sequence. Dosovitskiy \emph{et al.}~\cite{dosovitskiy2020image} introduced the Vision Transformer (ViT), which is a modified version of the standard transformer that takes the embedding sequences of image patches with ${k}$x${k}$ resolution as input~\cite{dosovitskiy2020image}. {Unlike the standard transformer in NLP, ViT discards the locality assumption and requires less vision-specific sensing bias, dominating in classification \cite{chen2021pre, bhojanapalli2021understanding, lanchantin2021general}
, semantic segmentation~\cite{seg2021, jin2021trseg, zheng2021rethinking}, object detection results~\cite{carion2020end, misra2021end, zhu2020deformable}, super-resolution restoration~\cite{liang2022light, zamir2022restormer}, depth estimation \cite{li2021revisiting, ding2022transmvsnet} and \emph{etc}. Chen \emph{et al.}~\cite{chen2021pre} proposed a multi-headed and multi-tailed within a shared backbone structure to cope with different vision tasks. Lanchantin \emph{et al.}~\cite{lanchantin2021general} proposed the classification transformer (C-Trans) network to complete a generic multi-label image classification task. Segmentation transformer (Segmenter)~\cite{seg2021} defined the semantic segmentation task as a sequence-to-sequence problem and employed the transformer architecture. Zheng \emph{et al.}~\cite{zheng2021rethinking} incorporated different decoders into ViT to tackle segmentation tasks. Detection transformer (DETR)~\cite{carion2020end} employed a transformer-based approach and treated object detection as a set prediction problem. Misra \emph{et al.}~\cite{misra2021end} added non-parametric queries and Fourier positional embeddings to the traditional transformer to suit the 3D object detection task. Zamir \emph{et al.}~\cite{zamir2022restormer} modified several key designs in multi-head attention and feed-forward network so that they can capture long-range pixel interactions while still being suitable for high-resolution images. Liang \emph{et al.}~\cite{liang2022light} introduced the ViT into light Light field image super-resolution restore tasks. Li \emph{et al.}~\cite{li2021revisiting} utilized dense pixel matching with location information and attention mechanisms in the transformer to take place of customer construction widely used for depth estimation. Ding \emph{et al.}~\cite{ding2022transmvsnet} designed a novel end-to-end deep neural network based on a feature matching transformer (FMT).
Despite RGB image fields, current works have also commenced scrutinizing the application of transformers in hyperspectral images (HSI) \cite{he2021spatial, qing2021improved, sun2022spectral, zhou2022multispectral} and achieved superior records. He \emph{et al.}~\cite{he2021spatial} introduced a new classification framework spatial-spectral transformer (SST) comprising an improved dense transformer layer for HSI classification. Sun \emph{et al.}~\cite{sun2022spectral} improved a spectral–spatial feature tokenization transformer (SSFTT) method to capture spectral–spatial features and high-level semantic features. Multispectral fusion transformer network (MFTNet)~\cite{zhou2022multispectral} designed a novel feature fusion tactic to generate robust cross-spectral fusion features.} 
Researchers have proposed a series of variants to improve the general capability of ViT. These variants contain substantial skillful tactics such as enhanced locality, improved self-attention algorithms, and structural redesign~\cite{chu2021conditional, li2021localvit, chen2021crossvit, liu2021swin, yang2021uncertainty}. 
{To introduce the locality principle in the transformer, Chu \emph{et al.}~\cite{chu2021conditional} proposed conditional positional vision transformer (CPVT), which uses a conditional positional encoding scheme consisting of a 2D CNN to realize translation invariance. Positional embeddings are generated based on the local relationship of the restricted tokens, which encode the relative location information of tokens implicitly~\cite{chu2021conditional}. Locality vision transformer (LocalViT)~\cite{li2021localvit} is inspired by the comparison between feed-forward networks (FFN) and reverse residual blocks, and depth-wise convolutional is applied to FFN to add locality to the vision transformer~\cite{li2021localvit}. Cross-scale attention transformer (CrossFormer)~\cite{W2021CrossFormer} presented multi-scale feature representation learning tactics in combination with a vision transformer. Cross-attention multi-scale vision transformer (CrossViT)~\cite{chen2021crossvit} proposed a two-branch transformer to process tokens generated by patches of different sizes and then fused these tokens multiple times to achieve mutual complementation of semantic information by applying cross-attention interaction~\cite{chen2021crossvit}. Liu \emph{et al.}~\cite{liu2021swin} proposed a hierarchical vision transformer using shift windows (swin-transformer), using a shift-window-based module to replace the traditional multi-head self-attention and the framework allows cross-window connection and promotes the flexibility of modeling at different scales. Considering the transformer's powerful global modeling capability and successful application in visual works, we design a transformer-based network further to explore its potential in the cross-view geo-localization task.}

\begin{table}[htbp]
\renewcommand\arraystretch{1.5}
\caption{List of \bf{Abbreviations}.}
    \begin{adjustwidth}{-\extralength}{0cm}
    \newcolumntype{C}{>{\centering\arraybackslash}X}
    \begin{tabularx}{\fulllength}{cCcC}
    \toprule
    {\bf{Abbreviation}} & {\bf {Explanation}} & {\bf{Abbreviation}} & {\bf {Explanation}}\\
    \midrule
    {CVGL}&{Cross-view geo-localization} &
    {MGTL}&{Mutual generative transformer learning}\\
    {CAMask}&{Cascaded attention masking}&
    {CVI}&{Cross-view interaction}\\
    {G2S}&{Ground-to-Satellite} &
    {VIFE}&{View independent feature extractor}\\ 
    {S2G}&{Satellite-to-Ground} &
    {SA}&{Spatial attention}\\
    {SCE}&{Spatial context enhancement} &
    {MSFA}&{Mutil-scale feature aggregation} \\
    {GKST}&{Generative knowledge supported transformer} & &\\
    \bottomrule
  \end{tabularx}%
  \end{adjustwidth}
  \label{tab:0}%
\end{table}%

\section{METHOD}\label{sec:3}
\subsection{Problem Formulation} {Let the cross-view geo-localization (CVGL) model be indicated as the function {$\mathcal F_{\Theta}$} parameterized by weights $\Theta$, which takes an image pair consists a ground-view image $\mathbf I_G$ and a satellite-view image $\mathbf I_S$ as input and produces their corresponding representations $\mathbf{F}_G$ and $\mathbf{F}_S$. Our goal is to learn $\Theta$ from the labeled training triplets $\{\mathbf{I}_G^i, \mathbf{I}_{SP}^i, \mathbf{I}_{SN}^i \}_{i=1}^N$ to make $\mathbf{F}_G$ and $\mathbf{F}_S$ closer while their corresponding cross-view images are matching, where $\mathbf{I}_G^i$ is the ground-view image, $\mathbf{I}_{SP}^i$ and $\mathbf{I}_{SN}^i$ are the positive and negative samples relative to $\mathbf{I}_G^i$, respectively. The process can be formulated as follows:}

\begin{equation}
\begin{array}{ll}
&{\mathbf{F}_G, \mathbf{F}_S=\mathcal F_{\Theta}(\mathbf{I}_G; \mathbf{I}_S)},\\ 
&{||\mathbf{I}_G^i - \mathbf{I}_{SP}^i||^{2} + \alpha < ||\mathbf{I}_G^i - \mathbf{I}_{SN}^i||^{2}}
\end{array}
\label{eq:0}
\end{equation}
where $\alpha$ is the margin in the triplet loss.

\subsection{View Independent Feature Extractor ($f_{\tt VIFE}$)}

\subsubsection{Overview}
\begin{figure}[t!]
\begin{adjustwidth}{-\extralength}{0cm}
\centering
\includegraphics[width=1.0\linewidth]{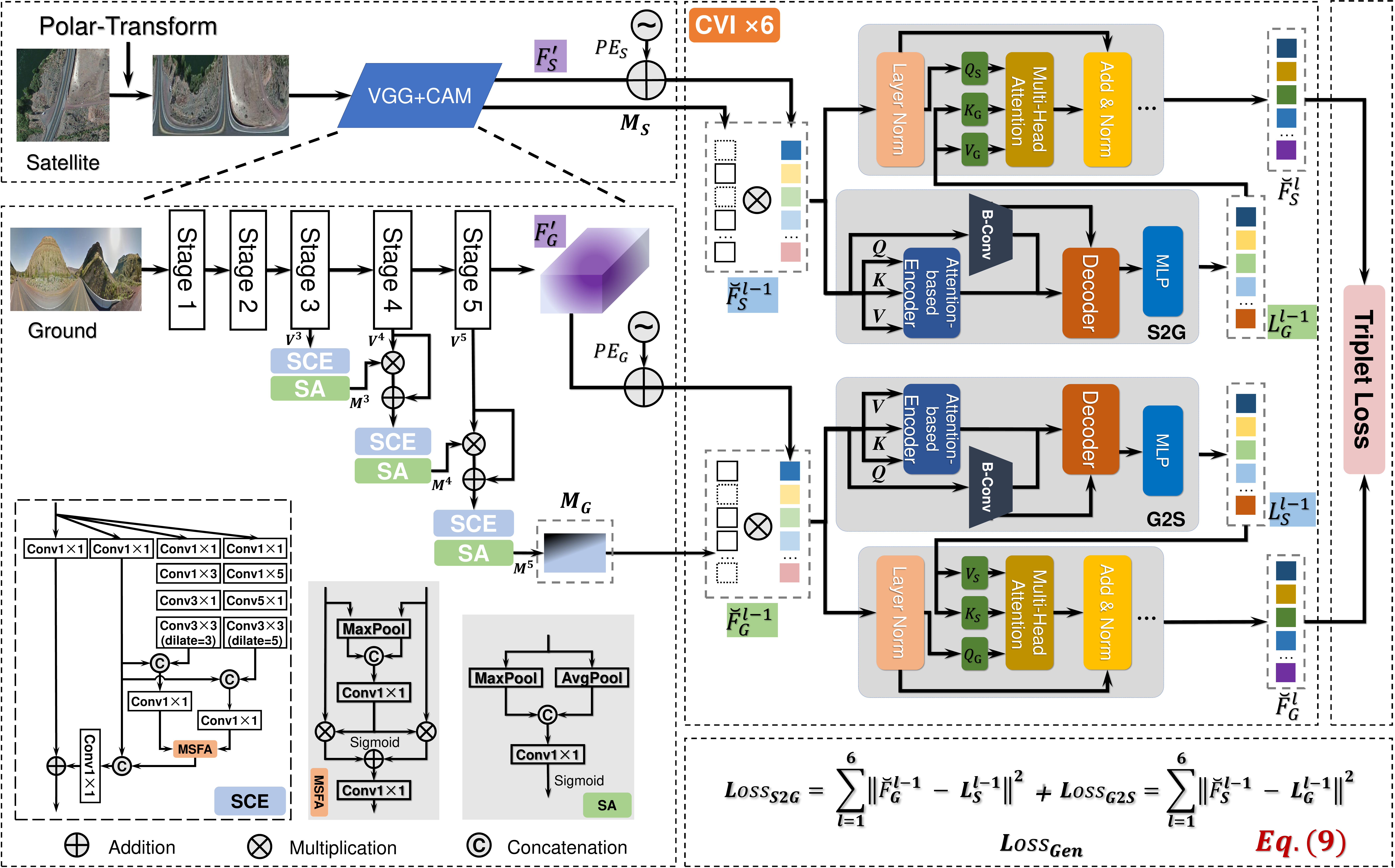}
\end{adjustwidth}
\caption{The overview of the proposed MGTL network.}\label{fig:3}
\end{figure}

{Here we retain the initial 13 convolutional layers in VGG16~\cite{simonyan2014very} and split them into 5 stages according to spatial resolutions, to extract high-order features from input images. Then we design a cascaded attention masking (CAMask) algorithm for learning fine-grained co-visual relationships by cascading multi-branch convolutional modules. Figure~\ref{fig:3} illustrates the overview of our proposed mutual generative transformer learning (MGTL). As mentioned above, $f_{\tt VIFE}$ takes an image pair \textless$\mathbf{I}_G$, $\mathbf{I}_S${\textgreater} as input, and produces two view-specified semantic representations \textless$\mathbf{F}^{'}_G$, $\mathbf{F}^{'}_S${\textgreater} and corresponding spatial attention masks \textless$\mathbf{M}_G$, $\mathbf{M}_S${\textgreater}, following Eq~\ref{eq:1} and Eq~\ref{eq:2}. Additionally, we list the main abbreviations in Table~\ref{tab:0} for simple reading.}

\subsubsection{Feature Extractor}
Formally, given an image pair $\mathbf{I}_G \in \mathbb{R}^{H_1\times W_1 \times 3}$ and $\mathbf{I}_S \in \mathbb{R}^{H_2\times W_2 \times 3}$, a multi-branch backbone (\emph{i.e.}, a Siamese-like VGG-based convolutional network with parameters $\Theta_{\tt VIFE}$) is performed to extract features and generate spatial attention masks for each view simultaneously:
\begin{equation}
\mathbf{F}^{'}_G=f_{\tt VIFE}(\mathbf{I}_G; \Theta_{\tt VIFE});
\mathbf{F}^{'}_S=f_{\tt VIFE}(\mathbf{I}_S; \Theta_{\tt VIFE})
\label{eq:1}
\end{equation}
where $\mathbf{F}^{'}_G \in \mathbb{R}^{c\times h \times w}$ and $\mathbf{F}^{'}_S \in \mathbb{R}^{c\times h \times w}$ are semantic representations with $c$ channels and $h \times w$ spatial resolutions for ground-view and satellite-view, respectively.

\subsubsection{Cascaded Attention Masking}
Viewpoint change results in drastic appearance differences, which means much redundant information exists in $\mathbf{F}^{'}_G$ and $\mathbf{F}^{'}_S$ while matching. To encourage the network to focus on the co-visual regions, we design a cascaded attention masking (CAMask) algorithm and integrate it into backbone VGG16~\cite{simonyan2014very}, seeking to learn the spatial attention masks to inhibit the non-co-visual areas adaptively. Figure~\ref{fig:3} (Left) illustrates the basic structure of the CAMask. Generally, CAMask takes the side-output features $\{\mathbf{V}^{i}\}_{i=3}^5$ generated by the backbone as input and produces the spatial attention masks $\mathbf{M}$ to enhance the inter-view co-visual information. \added[id=A., comment={Add more attention mask calculation details}]{Specifically, the fine-grained feature map captured by spatial context enhancement (SCE) is fed into two parallel pooling layers (i.e. maxpooling and avgpooling) along the channel dimension to generate two single channel feature maps, respectively. Subsequently, these feature maps are concatenated along the channel dimension, and a convolutional layer is employed to adaptively generate masks with $h \times w$ resolutions. Spatial Attention (SA) is illustrated in Figure~\ref{fig:3}.} Note that, for the expression brevity, $\mathbf{M}$ can refer to $\mathbf{M}_G$ or $\mathbf{M}_S$. The cascaded process can be formulated as follows:
\begin{equation}
\begin{array}{ll}
&\mathbf{M}^3 = \mathbf{SA}(\mathbf{SCE}(\mathbf{V}^3)),\\
&\mathbf{M}^4 = \mathbf{SA}(\mathbf{SCE}(\mathbf{V}^4 \otimes \mathbf{M}^3 + \mathbf{V}^4)),\\
&\mathbf{M}^5 = \mathbf{SA}(\mathbf{SCE}(\mathbf{V}^5 \otimes \mathbf{M}^4 + \mathbf{V}^5)),\\
\end{array}
\label{eq:2}
\end{equation}
where $\mathbf{M}^i$ represents the spatial attention mask of the $i$-th stage, $\mathbf{V}^i$ represents the feature map produced by the $i$-th stage in the backbone and $\mathbf{M}^5 \in \mathbb{R}^{h \times w}$ is the final spatial attention mask $\mathbf{M}$. A better understanding of CAMask can be gained by focusing on its two components: spatial context enhancement (SCE) and spatial attention (SA).

{\bf Spatial Context Enhancement (SCE).} \added[id=A., comment={Writing polishing.}]{To capture nuanced co-visual relationships, we meticulously devise a novel multi-branch convolutional module that effectively extracts fine-grained spatial representations from each view by utilizing diverse receptive fields.} Fan~\cite{fan2021concealed} proposed a texture-enhanced module (TEM) consisting of multiple convolutional branches with different receptive fields. There is evidence that it facilitates the sensitive capture of small spatial shifts. There are, however, certain limitations to coarse direct concatenation in TEM when the convolutional branches are independent. {Motivated by this, we design the SCE equipped with the multi-scale feature aggregation (MSFA) module to integrate branches with the guidance of spatial attention mechanism.} As shown in Figure~\ref{fig:3} (Left), the SCE includes a shortcut branch and three parallel residual branches $\{b_i\}_{i=1}^3$ with different dilation rates $d$ $\in$ \{1,3,5\}, respectively. The shortcut branch utilizes a 1 $\times$ 1 convolutional layer to generate $h_0$ with channel size $C$. The branch ${b}_1$ only contains a 1 $\times$ 1 convolutional layer to halve the channel, while the remaining two branches $\{b_i\}_{i=2}^3$ adopt 1 $\times$ 1 convolutional layer to reduce the channel  and consist of three convolutional layers, \emph{i.e.}, 1 $\times$ (2$i$ - 1) convolutional layer, (2$i$ - 1) $\times$ 1 convolutional layer, and 3 $\times$ 3 convolutional layers with dilation rate (2$i$ - 1), to fully explore the spatial context information with rich receptive fields. Let the $\{h_i\}_{i=1}^3$ represent the feature maps produced by the residual branches $\{b_i\}_{i=1}^3$, respectively. To fully explore the multi-scale information from the features $\{h_i\}_{i=1}^3$ generated by different convolutional layers, we carefully design a {\bf multi-scale feature aggregation (MSFA)} module by taking into account the specificities of spatial regions rather than concatenating them directly. Specifically, we concatenate the features $\{h_i\}_{i=2}^3$ with $h_1$ and fed the concatenated feature maps into a 1 $\times$ 1 convolutional layer to produce features $\{h_i^{'}\}_{i=2}^3$ with unified channel $C$. MSFA takes $\{h_i^{'}\}_{i=2}^3$ as input, and produces attention-aware feature maps $h_{msfa}$ which is then concatenated with $h_1$ followed by a 1 $\times$ 1 convolutional layer with a GeLU activation, and then added up with $h_0$ to produce the final enhanced contextual feature representation.

{\bf Spatial Attention (SA).} Inspired by~\cite{woo2018cbam}, we learn the spatial attention masks according to the enhanced contextual representations adaptively. In detail, SA takes the enhanced feature produced by SCE and eliminates the channel dimension by adopting the max and average pooling layers. In order to generate the spatial attention masks $\mathbf{M}_G (\mathbf{M}_S) \in \mathbb{R}^{h \times w}$, we concatenate the compact features obtained from the pooling layers and then apply a 1 $\times$ 1 convolutional layer with sigmoid activation.



To alleviate the limitation of feature location on the receptive learning field, we re-encode the features $\mathbf{F}^{'}_G$ and $\mathbf{F}^{'}_S$ with position information and enrich the co-visual areas by multiplying the spatial attention masks $\mathbf{M}_G$ and $\mathbf{M}_S$ generated by the cascaded attention masking (CAMask) algorithm:
\begin{equation}
\hat{\mathbf F}_G = (\mathbf{F}^{'}_G + \tt PE_G)\mathbf{M}_G;
\hat{\mathbf F}_S = (\mathbf{F}^{'}_S + \tt PE_S)\mathbf{M}_S
\label{eq:3}
\end{equation}
where $\hat{\mathbf F}_G$, $\hat{\mathbf F}_S \in \mathbb{R}^{l \times c}$ are compact and position-aware feature representations and $l=h \times w$. Following~\cite{dosovitskiy2020image}, $\tt PE_G$ and $\tt PE_S$ are the positional encoding of feature maps $\mathbf{F}^{'}_G$ and $\mathbf{F}^{'}_S$, respectively. 


\begin{figure}[t!]
\includegraphics[width=0.95\linewidth]{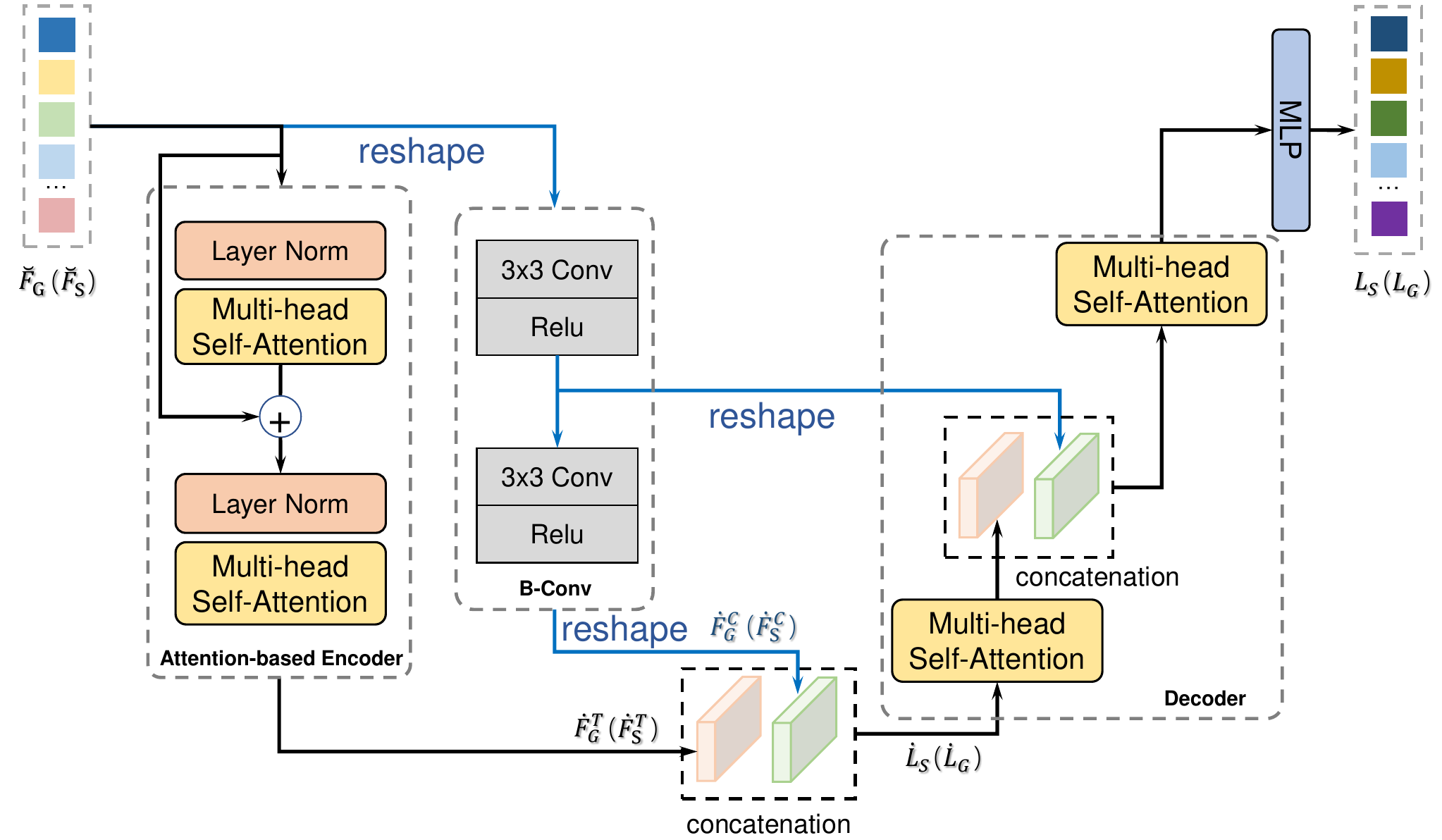}
\caption{The details of the cross-view generative module. The generative module is designed as an Unet-like~\cite{Ronneberger2015U} architecture, taking advantage of transformer and CNN features to extract contextual information.}
\label{fig:4}
\end{figure}
\unskip

\subsection{Cross-view Synthesis}

{A key principle of our proposed mutual generative transformer learning (MGTL) is cross-view interaction (CVI), which is achieved by generating mutual simulated knowledge through cross-view generative modules $f_{\tt G2S}$ and $f_{\tt S2G}$ with the supervision of generative loss in Eq~\ref{eq:9}. We emphasize that the ground view cannot obtain the matched satellite view in advance during the evaluation/localization period, which makes it impossible to directly take one view as input and produce the features of another view in the training phase. Therefore,  each generative module takes only the view feature from the self-branch as input to produce cross-view knowledge by using another view feature as supervision, which maintains that two sub-branches are completely decoupled while evaluating with unlabeled image pairs. Generative modules are embedded in transformer layers, and the generative knowledge is utilized to calculate the \emph{Key} and \emph{Value} while performing the attention mechanism. What's more, the generative module is trained with the whole transformer parts to fully mine semantic consistency across views using the generative knowledge-supported transformer, which we named generative transformer learning.} \added[id=A., comment={Writing polishing}]{Figure~\ref{fig:3} (Right) illustrates the overview of the proposed cross-view interaction (CVI), whereby one view's information is taken as input to generate knowledge that is aware of another view.} The co-visual enhanced and position-aware representations $\hat{\mathbf F}_G$ and $\hat{\mathbf F}_S$ are further normalized as $\breve{\mathbf F}_G={\tt LN}(\hat{\mathbf F}_G)$ and $\breve{\mathbf F}_S={\tt LN}(\hat{\mathbf F}_S)$ to maintain representational capacity, respectively, where ${\tt LN}$ indicates linear encoding operation following layer normalization. As shown in Figure~\ref{fig:3} (Right), the cross-view interaction module $f_{\tt CVI}$ is constructed by coupling two generative sub-modules $f_{\tt G2S}$ and $f_{\tt S2G}$ with an encoder-decoder structure as follows:
\begin{equation}
{\mathbf L}_S = f_{\tt G2S}(\breve{\mathbf F}_G),
{\mathbf L}_G = f_{\tt S2G}(\breve{\mathbf F}_S),
\label{eq:4}
\end{equation}

{\bf Cross-view Generative Module $f_{\tt G2S}$ and $f_{\tt S2G}$}. {Unet-like~\cite{Ronneberger2015U} architecture comprising encoder and decoder is widely used in generative tasks recently. Existing sduties~\cite{A2017Attention, wang2022transgcnn} demonstrate the attention mechanism in a transformer is excellent at modeling the global contextual information, and CNN excels at encoding local semantic information. With these properties in mind, we propose a novel generative module that owns Unet-like~\cite{Ronneberger2015U} architecture and combines multi-head self-attention and convolutional layers in parallel for mutual benefit.} Taking $f_{\tt G2S}$ as an example, the hidden feature representation $\breve{\mathbf F}_G$ is fed into the generative module to generate the simulated satellite-view feature representation ${\mathbf L}_G$, and the normalized satellite-view representation $\breve{\mathbf F}_S$ is used for supervision, and vice versa. {It's worth noting that both generative modules $f_{\tt G2S}$ and $f_{\tt S2G}$ own the same architecture but do not share weights due to the difference in input and generative content.}

\emph{\bf Encoder}: Figure~\ref{fig:4} illustrates the encoder-decoder architecture in detail. The encoder in the generative module is designed as a hybrid architecture that combines multi-head attention and convolutional layers. Taking $f_{\tt G2S}$ as an illustration, the feature ${\breve{\mathbf F}_G}$ is encoded independently by the attention layers and the convolutional layers, resulting in producing the compact features ${\Dot{\mathbf F}_G}^T$ and ${\Dot{\mathbf F}_G}^C$, respectively, and these two features are concatenated along the channel to form the encoded feature ${\Dot{\mathbf L}_S}$ which contains both global and local contextual information. \emph{\bf Decoder}: Following the acquisition of ${\Dot{\mathbf L}_S}$, the decoding process begins with a two-layer multi-head attention operation followed by multi-layer perceptions to generate simulated cross-view feature ${{\mathbf L}_S}$. The encoder and decoder are combined via skip connections to form an Unet-like~\cite{Ronneberger2015U} architecture, which enables aggregate features at different semantic levels.

\subsection{Generative Knowledge Supported Transformer (GKST) $f_{\tt GKST}$.}
Up till the present moment, we have acquired the inter-view representation $\breve{\mathbf F}_G$($\breve{\mathbf F}_S$) and the generative cross-view representation ${\mathbf L}_S$(${\mathbf L}_G$). To learn the final representation ${\mathbf F}_G$ and ${\mathbf F}_S$, we design a generative knowledge-supported transformer (GKST) to fully utilize all information. Formally, $f_{\tt GKST}$ takes $\breve{\mathbf F}_G \in \mathbb{R}^{l\times c}$($\breve{\mathbf F}_S \in \mathbb{R}^{l\times c}$) and ${\mathbf L}_S \in \mathbb{R}^{l\times c}$(${\mathbf L}_G \in \mathbb{R}^{l\times c}$) as inputs and produces the final high-order representations ${\mathbf F}_G$(${\mathbf F}_S$). Taking the ground view as an illustration, we feed the inter-view representation $\breve{\mathbf F}_G$ and cross-view knowledge ${\mathbf L}_S$ into a multi-head cross-attention layer to learn the cross-view enhanced features. The cross-attention process is formulated as follows:
\begin{equation}  \textstyle
\small
\begin{array}{ll}
&Q_G^i = \breve{\mathbf F}_G\mathbb{W}_Q^i, K_S^i={\mathbf L}_S\mathbb{W}_K^i, V_S^i={\mathbf L}_S\mathbb{W}_V^i,\\
&{\tt Head}_i = {\tt Attention}(Q_G^i, K_S^i, V_S^i), \\
&{\tt MH}(Q, K, V) = {\tt Concat}({\tt Head}_1, ..., {\tt Head}_n)\mathbb{W},
\end{array}
\label{eq:5}
\end{equation}
where $\mathbb{W}_Q^i$, $\mathbb{W}_K^i$, $\mathbb{W}_V^i$ and $\mathbb{W}$ are learnable parameters. The updated representations $\hat{\mathbf F}_G$ can be achieved by two residual connections, which are formulated as follows:
\begin{equation}
\begin{array}{ll}
&{\mathbf F}^{*}_G = {\tt MH}(Q, K, V) + \breve{\mathbf F}_G.,\\
&\hat{\mathbf F}_G = {\mathbf F}^{*}_G + {\tt LN}({\mathbf F}^{*}_G).
\end{array}
\label{eq:6}
\end{equation}
We can easily obtain the final satellite-view feature maps $\hat{\mathbf F}_S$ in a similar way.

{\bf Recurrent Learning Process.} To fully mine the benefits of the cross-view knowledge, we can further formulate the learning process recurrently as follows:
\begin{equation}  \textstyle
\small
\left\{ \begin{array}{ll}
&\hat{\mathbf F}^{l}_G = f_{\tt GKST} ({\mathbf L}^{l-1}_S, \breve{\mathbf F}^{l-1}_G), {\mathbf L}^{l-1}_S = f_{\tt G2S}(\breve{\mathbf F}^{l-1}_G), \\
&\hat{\mathbf F}^{l}_S = f_{\tt GKST} ({\mathbf L}^{l-1}_G, \breve{\mathbf F}^{l-1}_S), {\mathbf L}^{l-1}_G = f_{\tt S2G}(\breve{\mathbf F}^{l-1}_S),
\end{array}
\right.
\label{eq:7}
\end{equation}
where $\breve{\mathbf F}^{l-1}_G={\tt LN}(\hat{\mathbf F}^{l-1}_G)$, $\breve{\mathbf F}^{l-1}_S={\tt LN}(\hat{\mathbf F}^{l-1}_S)$. Note that, at the beginning ($l$=1), $\hat{\mathbf F}^{0}_G$ and $\hat{\mathbf F}^{0}_S$ are produced by Eq.~\ref{eq:3}, and the final representations ${\mathbf F}_G$ and ${\mathbf F}_S$ are produced by the last layer.

\subsection{Loss function}

Motivating the final representations $\mathbf{F}_G$ and $\mathbf{F}_S$ being more consistent between matching pairs but more discriminating among unmatching pairs. Following~\cite{shi2019spatial}, we employ a margin triplet loss $ {\mathbf Loss}_{\tt Triplet} $ for final representation supervision:
\begin{equation}
{\mathbf Loss}_{\tt Triplet}={\tt log}(1+e^{\gamma ({\tt d}_{pos}-{\tt d}_{neg})}),
\label{eq:8}
\end{equation}
where $\gamma$ indicates the function hyperparameter, ${\tt d}_{pos}$ and ${\tt d}_{neg}$ indicate the Euclidean distance between the positive and the negative pairs, respectively. To guarantee the quality of simulated cross-view knowledge. The cross-view knowledge generation module is supervised by mean squared errors (MSE) ${\mathbf Loss}_{\tt Gen}$:
\begin{equation}
{\mathbf Loss}_{\tt Gen} = \sum_{l=1}^{L}{( {||\breve{\mathbf F}^{l-1}_G - {\mathbf L}_S^{l-1}||}^{2} + {||\breve{\mathbf F}^{l-1}_S - {\mathbf L}_G^{l-1}||}^{2})},
\label{eq:9}
\end{equation}
where $\breve{\mathbf F}^{l-1}_G$($\breve{\mathbf F}^{l-1}_S$) and ${\mathbf L}^{l-1}_S$(${\mathbf L}^{l-1}_G$) denote inter-view representations and the generative cross-view knowledge at $l$-th recurrent step, respectively. $L$ is the total recurrent step. Finally, to learn the optimal parameters $\Theta$ for $\mathcal{F}_{\Theta}$, MGTL is jointly optimized through the overall learning ${\mathbf Loss}$, which is computed as:
\begin{equation}
{\mathbf Loss} = {\mathbf Loss}_{\tt Triplet} + \lambda {\mathbf Loss}_{\tt Gen},
\label{eq:10}
\end{equation}
where $\lambda$ is the balancing factor.


\section{EXPERIMENTS}\label{sec:4}
\subsection{Experimental Setting}
\begin{figure}[t!]
\centering
\includegraphics[width=13.5cm]{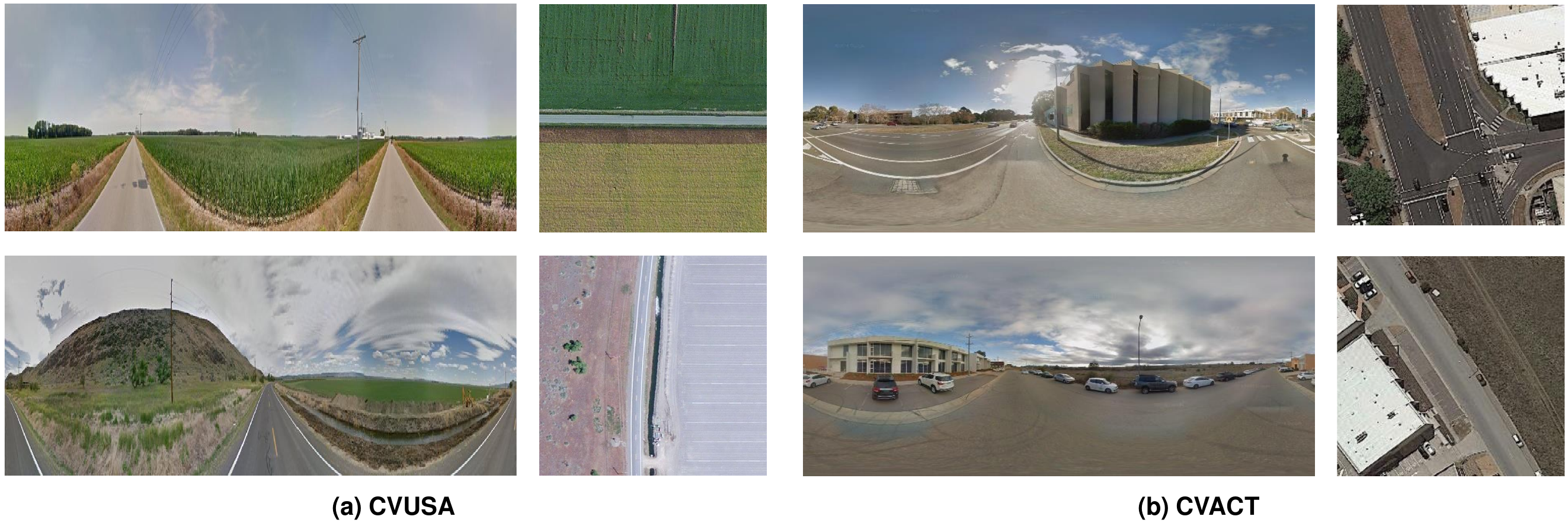}
\caption{Examples images pairs from \emph{CVUSA}~\cite{workman2015wide} and \emph{CVACT}~\cite{zhu2021geographic}.
\label{fig:5}}
\vspace{-2.5ex}
\end{figure}

{\bf Dataset:} Following~\cite{zhu2021vigor,shi2019spatial,yang2021cross}, we evaluate the performance of mutual generative transformer learning (MGTL) on two widely-used challenging benchmarks \emph{i.e., CVUSA}~\cite{workman2015wide} and \emph{CVACT}~\cite{liu2019lending}. \emph{CVUSA} was constructed by Workman \emph{et al.}~\cite{workman2015location} containing 1.7 million training pairs collected from San Francisco. {However, the relatively limited acquisition locations result in poor generalization capability of the extracted features when images from other positions are taken as input. To address this issue, the researchers reconstructed a new extensive dataset named \emph{CVUSA}~\cite{workman2015wide}, which contains 1.5 million geo-tagged pairs of ground-view and satellite-view images covering the continental United States, with resolutions of 1232x224 and 750x750, respectively. Ground-view images were collected using the Google Street View App and Flickr with different pre-processing methods. Specifically, The researchers randomly sampled images from the continental United States using the former but they divided the entire area into a 100x100 grid and sampled up to 150 images in each cell while using the latter. Further, based on the origin \emph{CVUSA}~\cite{workman2015wide}, Zhai \emph{et al.}~\cite{zhai2017predicting} selected ground-view panoramas from \emph{CVUSA}~\cite{workman2015wide} and satellite-view images from Bing Maps at the same location as matching pairs. Especially, the panoramas were wrapped to align with the satellite images using camera parameters.} Finally, they released a subset of the CVUSA~\cite{workman2015wide} containing 44416 ground-satellite image pairs collected at the same location in which 35532 training pairs and 8884 evaluation pairs. This subset becomes a widely-used benchmark because of its high resolution and simple format. To better investigate the possibility of matching geolocation in urban scenarios, Liu \emph{et al.}~\cite{liu2019lending} created a city-scale cross-view dataset \emph{CVACT}~\cite{liu2019lending} densely covering Canberra, Australia. Keeping the same with \emph{CVUSA}, ground-view panoramas were collected from Google Street View App at zoom $2$ with 1664x832 image resolution, while satellite-view images were collected from Google Map App at zoom 20 at the same location with 1200x1200 resolution. Especially, to fully evaluate the generalization of CVGL methods, \emph{CVACT}~\cite{liu2019lending} released \emph{CVACT}\_test containing extra 92802 challenging pairs for testing only. Figure~\ref{fig:5} displays several ground-satellite image pairs from \emph{CVUSA}~\cite{workman2015wide} and \emph{CVACT}~\cite{liu2019lending}.

{\bf Evaluation Metric:} Following the existing works ~\cite{shi2019spatial, yang2021cross}, the recall accuracy at top $K$(r@$K$) is performed to evaluate the proposed MGTL. Staying in step with these existing methods, $K=1, 5, 10, 1\%$ are selected.

{\bf Training Setting:}
During the training phase, MGTL adopts VGG16~\cite{simonyan2014very} pre-trained on ImageNet~\cite{deng2009imagenet} as the backbone. {All training images are resized to $112 \times 616$ resolutions augmented by random cropping, flipping, and rotation, \emph{etc.}. We employ Adam optimizer to optimize the whole network with the initial learning rate of $10^{-5}$. We set the recurrent learning step of the generative knowledge supported transformer (GKST) to 6 and equip 6 attention heads for each step. We set the batch size to 16 and train the network for up to 150 epochs until complete convergence. The balancing factor $\lambda$ in Eq.~\ref{eq:10} is carefully set to 0.05, and following~\cite{shi2019spatial}, the regular item $\gamma$ in Eq.~\ref{eq:8} is set to 10.0.}

{\bf Reproducibility:} We implement the MGTL based on TensorFlow and train the whole network on an NVIDIA GTX Titan X GPU with 12G CUDA memory.

\subsection{Main Results}
{\bf Baselines:} {Cross-view geo-localization (CVGL) has garnered significant research interest, resulting in several impressive works emerging in the field. To demonstrate the superiority of our proposed method, we select 17 strong baselines and state-of-the-art methods in total, i.e., Workman, \emph{et al.}~\cite{workman2015location}, Vo, \emph{et al.}~\cite{vo2016localizing}, Zhai, \emph{et al.}~\cite{zhai2017predicting}, Cross-view Matching Network (CVM-Net)~\cite{hu2018cvm}, Liu, \emph{et al.}~\cite{liu2019lending}, Regmi, \emph{et al.}~\cite{regmi2019bridging}, Spatial-aware feature aggregation network (SAFA)~\cite{shi2019spatial}, l Cross-View Feature Transport technique (CVFT)~\cite{shi2020optimal}, Dynamic Similarity Matching network (DSM)~\cite{shi2020looking}, Toker,~\emph{et al.}~\cite{toker2021coming}, Layer-to-Layer Transformer (L2LTR)~\cite{yang2021cross}, Local Pattern Network (LPN)~\cite{2021Each}, Unit SAFA+Subtraction Attention Module (USAM)~\cite{Lin2022usam}, LPN+USAM~\cite{Lin2022usam}, pure Transformer-based geo-localization (TransGeo)~\cite{zhu2022transgeo}, Transformer-guided Convolutional Neural
Network (TransGCNN)~\cite{wang2022transgcnn}, LPN+Dynamic Weighted Decorrelation Regularization (DWDR)~\cite{wang2022learning}.} Especially, for omnidirectional comparison, we use their recommended settings for training. Our MGTL outperforms existing methods across most top $K$(r@$K$) metrics on both benchmarks, showcasing the effectiveness of our proposed cascaded attention masking (CAMask) algorithm and cross-view interaction (CVI) tactic. In this section, we provide a detailed introduction to our experiment setup and experiment results.

\begin{figure}[htp]
\centering
\includegraphics[width=13.5cm]{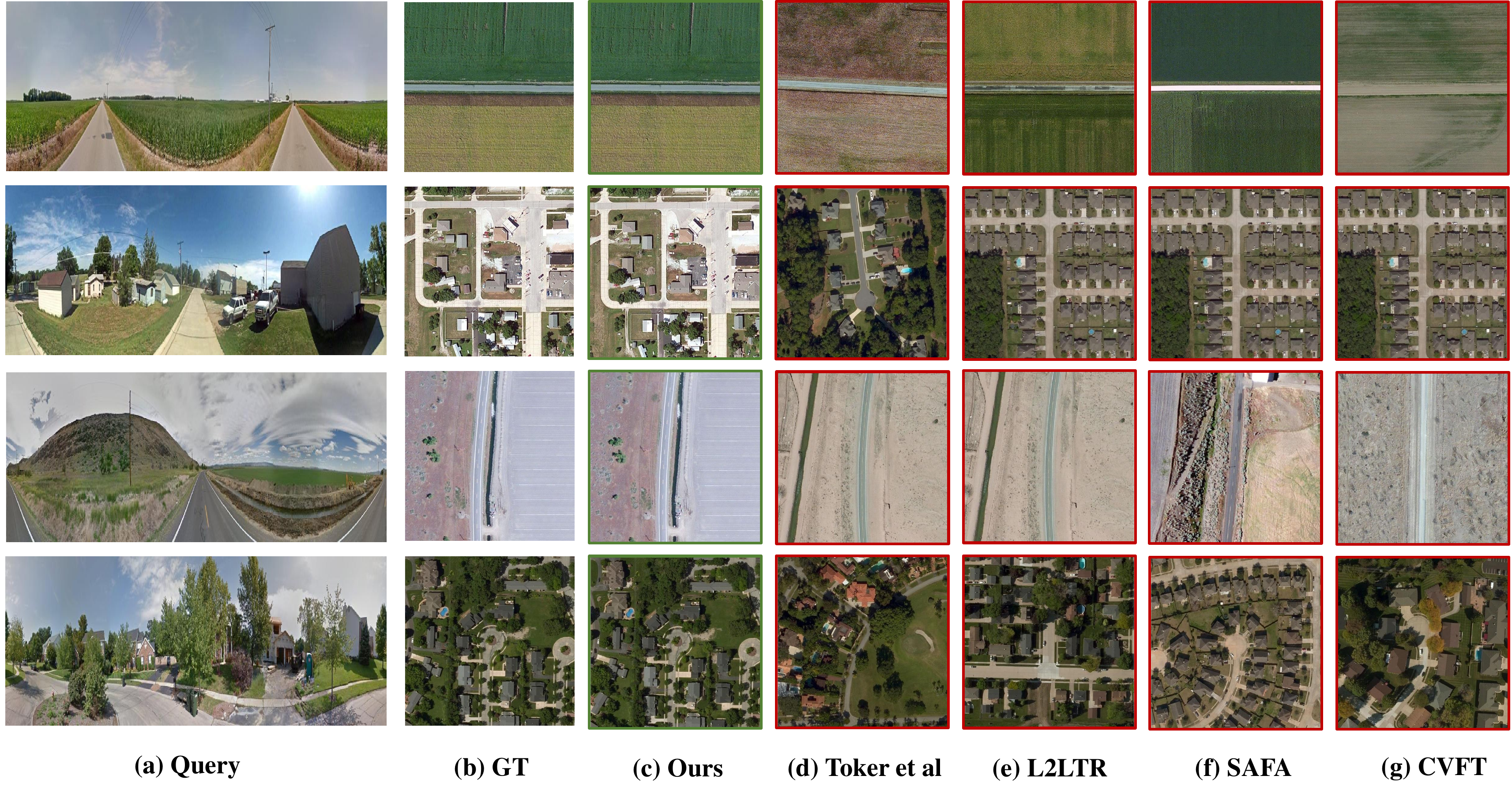}
\caption{Comparison results of some hard pairs.
\label{fig:6}}
\end{figure} 

{\bf Performance on \emph{CVUSA}:}
{The test set of \emph{CVUSA}~\cite{workman2015wide} has 8884 challenging ground-satellite image pairs. 
The results with 17 SOTAs presented in Table~\ref{tab:1} (Left) show that our approach achieves state-of-the-art performance compared to all baselines in terms of almost all top $K$(r@$K$) metrics on \emph{CVUSA}~\cite{workman2015wide}. Our approach achieves the best top 1(r@1) retrieval accuracy, significant increases of 4.34\% (90.16\% $\rightarrow$ 94.50\%) and 3.28\% (91.22\% $\rightarrow$ 94.50\%) over SAFA+USAM~\cite{Lin2022usam} and LPN+USAM~\cite{Lin2022usam}. Notably, although the top 1\% (r@$1$\%) retrieval accuracy is almost 100\%, MGTL still achieves 0.11\% growth.} Our approach outperforms L2LTR~\cite{yang2021cross} by 0.45\% (94.05\% $\rightarrow$ 94.50\%) in the top 1(r@$1$) retrieval accuracy while keeping less computation complexity and model capacity. TransGeo~\cite{zhu2022transgeo} utilizes a three-branch vision transformer~\cite{dosovitskiy2020image} with a novel attention-based masking scheme. Our results outperform it by 0.42\% (94.08\% $\rightarrow$ 94.50\%) in top 1(r@$1$) retrieval accuracy. Nevertheless, both methods except MGTL ignore the semantic consistency revealed by cross-view interaction, making mutual generative learning the more convincing method. Figure~\ref{fig:6} shows partial hard images pairs retrieval result from Toker \emph{et al.}~\cite{toker2021coming}, L2LTR~\cite{yang2021cross}, SAFA~\cite{shi2019spatial}, CVFT~\cite{shi2020optimal}. The similarity between the ground truth and the selected unmatched satellite images heavily interferes with other models. In contrast, this illustrates that the co-visual enhanced features learned by CAMask and CVI own finer-grained understandings of scenarios and are highly discriminative.

{\bf Performance on \emph{CVACT}\_val:}
The evaluation set of \emph{CVACT}~\cite{liu2019lending} contains 8884 ground-satellite image pairs, keeping consistent with \emph{CVUSA}~\cite{workman2015wide}. Table~\ref{tab:1} (Middle) presents the results with 13 STOAs on \emph{CVACT}\_val~\cite{liu2019lending}. Our approach achieves the best performance across all top K(r@K) metrics (r@1, r@5, r@10, r@1\%) on \emph{CVACT}\_val, \emph{i.e.}, 85.42\%, 94.64\%, 96.11\% and 98.51\%, respectively. MGTL achieves significant improvements over LPN+USAM\cite{Lin2022usam}, SAFA+USAM~\cite{Lin2022usam} and LPN+DWDR~\cite{wang2022learning}, with increasing top 1(r@1) retrieval accuracy of 3.40\% (82.02\% $\rightarrow$ 85.42\%), 3.02\% (82.40\% $\rightarrow$ 85.42\%) and 1.69\% (83.73\% $\rightarrow$ 85.42\%). In addition, we promote significantly across all metrics compared to classical Siamese-like VGG-based convolutional methods, \emph{e.g.}, SAFA~\cite{shi2019spatial}, CVFT~\cite{shi2020optimal}, and DSM~\cite{shi2020looking}. 
{The experiment results mentioned above demonstrate the effectiveness of our CAMask and CVI introduced by MGTL. In comparison with traditional transformer-based methods TransGeo~\cite{zhu2022transgeo} and L2LTR~\cite{yang2021cross}, MGTL increases significantly by 0.47\% (84.95\% $\rightarrow$ 85.42\%) and 0.53\% (84.89\% $\rightarrow$ 85.42\%) in top 1 (r@1) retrieval accuracy, respectively. This strongly proves the superiority of our generative knowledge-supported transformer framework.} Toker \emph{et al.}~\cite{toker2021coming} proposed a GAN-based method to synthesize realistic ground-view images from satellite images, which explores the benefits of generative learning for cross-view matching. MGTL outperforms it by 2.14\% in top 1(r@1) retrieval accuracy, showcasing that our mutual generative learning strategy is more effective and has extreme generalizability in urban scenarios. 

\begin{table}[t!]
\caption{\textbf{Quantitative results} on the \emph{CVUSA}~\cite{workman2015wide} and \emph{CVACT}~\cite{liu2019lending} dataset. }\label{tab:1}
    \begin{adjustwidth}{-\extralength}{0cm}
    \newcolumntype{C}{>{\centering\arraybackslash}X}
    \begin{tabularx}{\fulllength}{Ccccccccccccc}
        \toprule
        \multirow{2}{*}{\tabincell{c}{\textbf{Model}}} &
        \multicolumn{4}{c}{\tabincell{c}{\textbf{CVUSA}}} &
        \multicolumn{4}{c}{\tabincell{c}{\textbf{CVACT\_val}}} &
        \multicolumn{4}{c}{\tabincell{c}{\textbf{CVACT\_test}}} \\
         &  {r@1} & {r@5} & {r@10} & {r@1\%} & {r@1} & {r@5} & {r@10} & {r@1\%} & {r@1} & {r@5} & {r@10} & {r@1\%} \\
        \midrule
        \makecell[l]{2015~Workman, \emph{et al.}~\cite{workman2015location}}&  - & -     & -     & 34.30  & -     & -     & -     & - & -     & -     & -     & -\\
        \makecell[l]{2016~Vo, \emph{et al.}~\cite{vo2016localizing}} & -  & -     & -     & 63.70  & -     & -     & -     & - & -     & -     & -     & -\\
        \makecell[l]{2017~Zhai, \emph{et al.}~\cite{zhai2017predicting}} & -     & -     & -     & 43.20  & -     & -     & -     & - & -     & -     & -     & -\\
        \makecell[l]{2018~CVM-Net~\cite{hu2018cvm}} &  22.47 & 49.98 & 63.18& 93.62 & 20.15 & 45.00 & 56.87 & 87.57 & 5.41 &14.79&25.63&54.53  \\
        \makecell[l]{2019~Liu, \emph{et al.}~\cite{liu2019lending}} &40.79  & 66.82  & 76.36  & 96.12  & 46.96  & 68.28  & 75.48  & 92.04& 19.9  & 34.82  & 41.23  & 63.79 \\
        \makecell[l]{2019~Regmi, \emph{et al.}~\cite{regmi2019bridging}} &48.75  & -     & 81.27  & 95.98  & -     & -     & -     & - & -     & -     & -     & -\\ \makecell[l]{2019~SAFA~\cite{shi2019spatial}}&89.84&96.93&98.14&99.64&81.03&92.80&94.84&98.17&55.50&79.94&85.08&94.49 \\
        \makecell[l]{2020~CVFT~\cite{shi2020optimal}}&61.43&84.69&90.49&99.02&61.05&81.33&86.52&95.93&34.39&58.83&66.78&95.99 \\
        \makecell[l]{2020~DSM~\cite{shi2020looking}}&91.96&97.50&98.54&99.67&82.49&92.44&93.99&97.32&35.55&60.17&67.95&86.71 \\ \makecell[l]{2021~Toker,~\emph{et al.}~\cite{toker2021coming}}&92.56&97.55&98.33&99.57&83.28&93.57&95.42&98.22&61.29&85.13&89.14&98.32 \\
        \makecell[l]{2021~L2LTR~\cite{yang2021cross}}&94.05&98.27&98.99&99.67&84.89&94.59&95.96&98.37&60.72&85.85&89.88&96.12\\
        \makecell[l]
        {2021~LPN~\cite{2021Each}}&93.78&98.50&99.03&99.72&82.87&92.26&94.09&97.77&-&-&-&-\\
        \makecell[l]
        {2022~SAFA+USAM~\cite{Lin2022usam}} &90.16 &- &- &99.67 &82.40 &- &- &98.00 &56.16 &- &- &95.22 \\
        \makecell[l]
        {2022~LPN+USAM~\cite{Lin2022usam}} &91.22 &- &- &99.67 &82.02 &- &- &98.18 &37.71 &- &- &87.04 \\
        \makecell[l]
        {2022~TransGeo~\cite{zhu2022transgeo}}  &94.08 &98.36 &99.04 &99.77 &84.95 &94.14 &95.78 &98.37 &- &- &- &- \\
        \makecell[l]
        {2022~TransGCNN~\cite{wang2022transgcnn}}  &94.15 &98.21 &98.94 &99.79 &84.92 &94.46 &95.88 &98.36 &- &- &- &- \\
        \makecell[l]
        {2022~LPN+DWDR~\cite{wang2022learning}}  &94.33 &\bf98.54 &99.09 &\bf99.80 &83.73 &92.78&94.53 &97.78 &- &- &- &- \\
        \makecell[l]{\textbf{Ours}} &\bf94.50  &98.41  &\bf99.20  &99.78 &\bf85.42  &\bf94.64  &\bf96.11  &\bf98.51&\bf61.55&\bf86.61&\bf90.74&{\bf98.46}\textsuperscript{1}\\
        \bottomrule
    \end{tabularx}
    \end{adjustwidth}
    \noindent{\footnotesize{\textsuperscript{1} Results are cited directly, and the best results are highlighted.}}
\end{table}



{\bf Performance on \emph{CVACT}\_test:}
\emph{CVACT}\_test is extremely massive and challenging, consisting of 92802 ground-satellite image pairs in urban scenarios for testing only. For the challenging \emph{CVACT}\_test, we compare our approach with 9 SOTAs. As shown in Table~\ref{tab:1} (Right), our MGTL sets new retrieval accuracy records across all metrics compared to existing SOTAs. MGTL increases the top 1(r@1) retrieval accuracy significantly by 0.83\% (60.72\% $\rightarrow$ 61.55\%) and 5.39\% (56.16\% $\rightarrow$ 61.55\%) compared to L2LTR~\cite{yang2021cross} and SAFA+USAM~\cite{Lin2022usam}, respectively. 
{Furthermore, our results not only outperform others in top 1(r@1) retrieval accuracy, but also gain a remarkable increase of 1.48\% (85.13\% $\rightarrow$ 86.61\%) in top 5(r@5) retrieval accuracy over Toker \emph{et al.}~\cite{toker2021coming} and 2.34\% (96.12\% $\rightarrow$ 98.46\%) in top 1\%(r@1\%) recall accuracy over L2LTR~\cite{yang2021cross}.}
Superior experiment results showcase that MGTL is capable of capturing high-order understandings of cross-view scenarios essential for CVGL in unfamiliar environments absence of prior knowledge.


\subsection{Ablation Study}
As mutual generative transformer learning (MGTL) incorporates the cascaded attention masking (CAMask) algorithm and cross-view interaction (CVI) tactic into the cross-view geo-localization (CVGL) task, we conduct substantial ablation studies to carefully scrutinize how each component affects the learning capability of the model.

\begin{figure}[t!]
\begin{adjustwidth}{-\extralength}{0.25cm}
\centering
\includegraphics[width=0.8\linewidth]{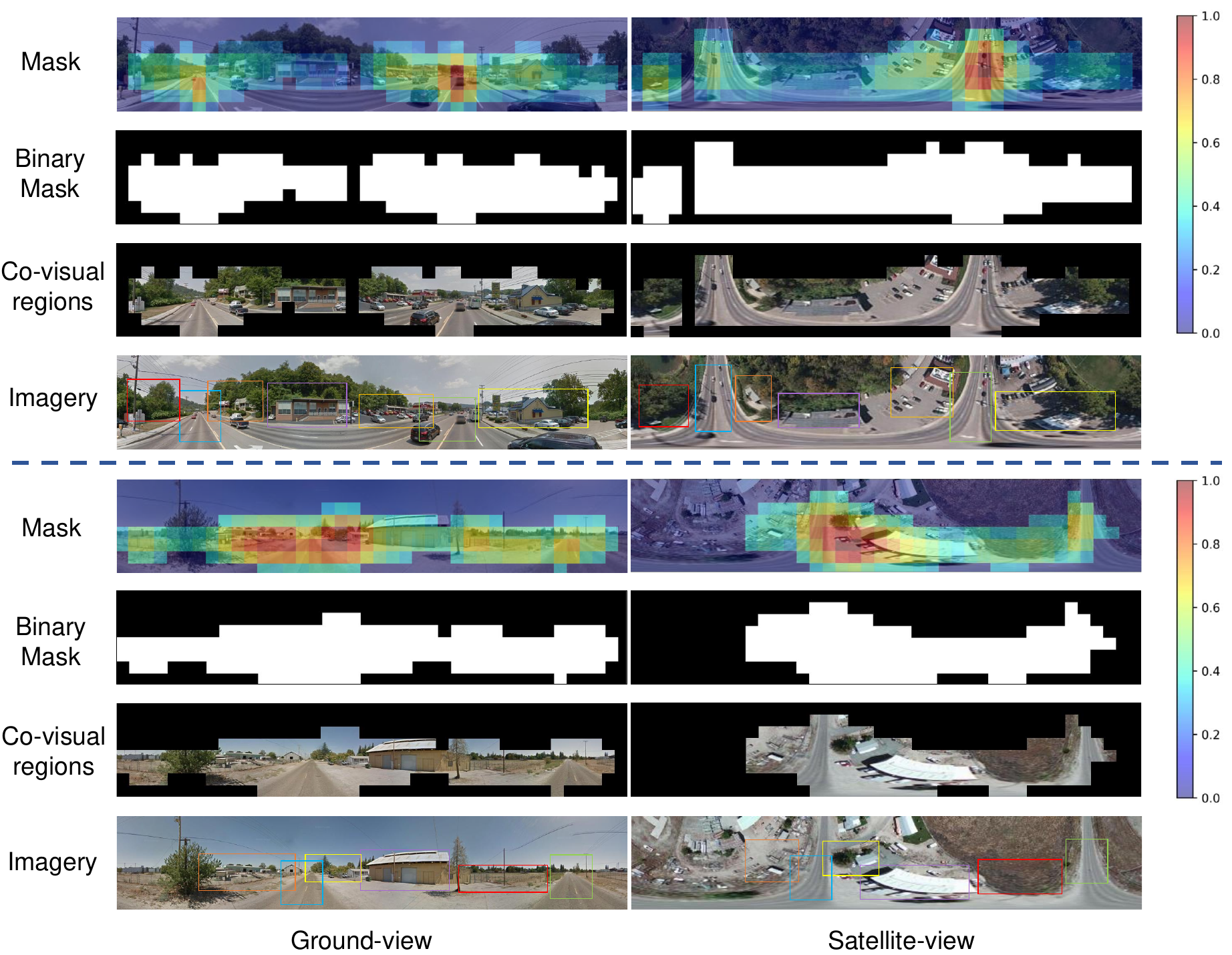}
\end{adjustwidth}
\caption{Visualization results of cascaded attention masks.}
\label{fig:7}
\end{figure}

{\bf Effectiveness of CAMask:} To qualitatively study the effectiveness of our proposed CAMask algorithm, we inspect the performance of the backbone VGG16~\cite{simonyan2014very} with fully-connected layers removed. As shown in Table~\ref{tab:2}, all metrics degrade significantly while removing CAMask. The top 1(r@1) retrieval accuracy suffers a drastic decrease by 10.18\% from 90.12\% to 79.94\%, supporting the notion that co-visual information explicitly learned by CAMask is extremely critical for CVGL. {In Addition, we remove the CAMask from the whole equipped model. Observing the last two lines in Table~\ref{tab:2}, the top 1(r@1) retrieval accuracy still suffers a heavy decrease by 4.06\% from 94.50\% to 90.44\%, suggesting once again that the above notion is strongly convinced.}
To explore the necessity of SCE and SA, Table~\ref{tab:3} displays the comparison results while removing any of them, respectively. Once we replace the SCE with fully convolutional blocks, the top 1(r@1) retrieve accuracy decreases by 2.21\% (94.50\% $\rightarrow$ 92.29\%) and 3.05\% (85.42\% $\rightarrow$ 82.37\%) on \emph{CVUSA} and \emph{CVACT\_val}, respectively. Similarly, we replace the SA with global average pooling (GAP), the top 1(r@1) retrieval accuracy degrades by 0.88\% (94.50\% $\rightarrow$ 93.62\%) and 1.09\% (85.42\% $\rightarrow$ 84.33\%) on both two datasets. MGTL suffers precision decreases to varying content, suggesting co-visual enhanced feature representations learned by CAMask lead to more reliable results. 
{To show the superiority of CAMask qualitatively, we meticulously visualize the cascaded attention masks in Figure~\ref{fig:7} to support our claim. The first row indicates the generative attention masks and as well as corresponding attention scores. To showcase the co-visual regions intuitively, we binarize the attention masks, as shown in the second row. Subsequently, the original images are cropped with the guidance of binary masks, as shown in the third row. Observing the third row, only the co-visual regions (\emph{e.g.} road, building) remain, and redundant non-co-visual regions (\emph{e.g.} 'sky' in ground imagery but absent in satellite imagery) useless for matching are masked. CAMask eradicates these disturbances in a simple but effective manner. Finally, to showcase the correctness of co-visual relationships, the same regions captured across views are marked with rectangles of the same color, as shown in the fourth row.}

\begin{table}[htb]
\caption{\textbf{Ablation study} of the proposed cascaded attention masking (CAMask) algorithm.} 
\label{tab:2}
    \begin{adjustwidth}{-\extralength}{0cm}

    \newcolumntype{C}{>{\centering\arraybackslash}X}
    \begin{tabularx}{\fulllength}{CCCCCcccccccc}
        \toprule
        \multicolumn{3}{c}{\tabincell{c}{\textbf{Candidate}}} & 
        \multicolumn{2}{c}{\tabincell{c}{\textbf{Complexity}}} &
        \multicolumn{4}{c}{\tabincell{c}{\textbf{CVUSA}}} & 
        \multicolumn{4}{c}{\tabincell{c}{\textbf{CVACT\_val}}}\\
        VGG16 & CAMask & CVI & GFLOPs $\downarrow$ & Param.$\downarrow$ & {r@1} & {r@5} & {r@10} & {r@1\%} & {r@1} & {r@5} & {r@10} & {r@1\%} \\
        \midrule
        \ding{52}  &          & & 28.22 &29.42M &79.94  &93.66 &96.25 &99.31 &70.67 &87.73 &91.13 &95.78 \\
        \ding{52}  &\ding{52} & & 59.02 & 91.18M &85.15  &95.13 &96.89 &99.43 &76.32 &89.64 &92.26 &96.21  \\
        \ding{52}  & & \ding{52} &  28.90  &  137.21M &  90.44  &  96.83 &  97.41 &  99.45 &  81.25 &  92.12 &  94.38 &  97.69  \\
         \ding{52} &  \ding{52} &  \ding{52} &  59.71 &  171.97M & \bf94.50  &  \bf98.41  &  \bf99.20  &  \bf99.78  &  \bf85.42  &  \bf94.64  &  \bf96.11  &  {\bf98.51}\\
        \bottomrule
    \end{tabularx}
    \end{adjustwidth}
\end{table}

\begin{table}[t!]
    \caption{\textbf{Ablation study} of the proposed spatial attention (SA) and spatial context enhancement (SCE) in cascaded attention masking (CAMask).} 
    \begin{adjustwidth}{-\extralength}{0cm}
    \newcolumntype{C}{>{\centering\arraybackslash}X}
    \begin{tabularx}{\fulllength}{CCCcccccccccc}
        \toprule
        \multicolumn{3}{c}{\tabincell{c}{\textbf{Candidate}}} &
        \multicolumn{2}{c}{\tabincell{c}{\textbf{Complexity}}} &
        \multicolumn{4}{c}{\tabincell{c}{\textbf{CVUSA}}} & 
        \multicolumn{4}{c}{\tabincell{c}{\textbf{CVACT\_val}}}\\
        VGG16+CVI & w/SA & w/SCE & GFLOPs $\downarrow$ &Param.$\downarrow$ &{r@1} & {r@5} & {r@10} & {r@1\%} & {r@1} & {r@5} & {r@10} & {r@1\%} \\
        \midrule
         \ding{52}  & & &  28.90  &  137.21M &  90.44  &  96.83 &  97.41 &  99.45 &  81.25 &  92.12 &  94.38 &  97.69  \\
        
        \ding{52} &\ding{52}  &            &28.90 &137.21M &92.29  &97.65 &98.67&99.72  &82.37 &93.35&95.17  &98.23  \\
        \ding{52} &           &\ding{52}   &59.71 &171.97M &93.62  &98.41  &99.07  &99.73  &84.33  &94.31  &95.67 &98.43   \\
        \ding{52} &\ding{52}  &\ding{52}  &59.71 & 171.97M &\bf94.50  &\bf98.41  &\bf99.20  &\bf99.78  &\bf85.42  &\bf94.64  &\bf96.11  &{\bf98.51}\textsuperscript{1}\\

        \hline
    \end{tabularx}
    \end{adjustwidth}
    \label{tab:3}
    \noindent{\footnotesize{\textsuperscript{1} 'w/' and 'w/o' means the proposed MGTL is equipped with SA or SCE, respectively.}}
\end{table}

{\bf Effectiveness of CVI:} We introduce the CVI tactic to implicitly explore co-visual information and empirically investigate the superiority of CVI in MGTL. Table~\ref{tab:4} shows that when incorporating transformer~\cite{A2017Attention} blocks, the top 1(r@1) retrieval accuracy increases by 6.27\% (85.15\% $\rightarrow$ 91.42\%). However, this is still a sub-optimal performance compared to existing SOTAs. {'w/o' in Table~\ref{tab:4} refers to the pure transformer with the absence of CVI, all metrics suffer drastic degradations, in which top 1(r@1) retrieval accuracy decreases by 3.08\% (94.50\% $\rightarrow$ 91.42\%), showcasing that CVI further boosts the pure transformer learning  capability and enhance the similarity of feature representations between matching pairs.}

\begin{table}[t!]
\caption{{\bf Ablation study} of cross-view interaction (CVI).}
    \begin{adjustwidth}{-\extralength}{0cm}
    \newcolumntype{C}{>{\centering\arraybackslash}X}
    \begin{tabularx}{\fulllength}{CCCcccccccccc}
        \toprule
        \multicolumn{3}{c}{\tabincell{c}{\textbf{Candidate}}} &
        \multicolumn{2}{c}{\tabincell{c}{\textbf{Complexity}}} &
        \multicolumn{4}{c}{\tabincell{c}{\textbf{CVUSA}}} & \multicolumn{4}{c}{\tabincell{c}{\textbf{CVACT\_val}}}\\
         VGG16+CAMask & w/o & w/CVI &GFLOPs$\downarrow$ &Param.$\downarrow$ & {r@1} & {r@5} & {r@10} & {r@1\%} & {r@1} & {r@5} & {r@10} & {r@1\%} \\
         \midrule
         \ding{52}& &  &59.02 &91.98M &85.15  &95.13 &96.89 &99.43 &76.32 &89.64 &92.26 &96.21 \\

         \ding{52}&\ding{52} & & 59.32&113.48M &91.42 &96.21 &98.04 &99.62 &81.99 &93.16 &95.04 &98.23  \\
         \ding{52}& &\ding{52} & 59.71&171.97M &\bf94.50  &\bf98.41  &\bf99.20  &\bf99.78&\bf85.42  &\bf94.64  &\bf96.11  &{\bf98.51} \textsuperscript{1}\\
         \bottomrule
    \end{tabularx}
    \end{adjustwidth}  
    \label{tab:4}
    \noindent{\footnotesize{\textsuperscript{1}'w/' and 'w/o' means the transformer learning is or is not equipped with CVI, respectively.}}
    
\end{table}

{\bf Composition of Generative Module:} 
To determine the most effective interaction mode, we empirically explore the Variational Autoencoder (VAE)~\cite{kingma2013auto}, CNN-based Unet~\cite{Ronneberger2015U}, and pure transformer~\cite{A2017Attention} block. \added[id=A., comment={Writing polishing.}]{The results shown in Table~\ref{tab:5} suggest the superiority of our hybrid generative module. Specifically, VAE~\cite{kingma2013auto} is considered as one of the classical generative models, wherein we utilize 2 two-layer fully-connected networks as encoder and decoder, respectively. Follow~{\cite{Ronneberger2015U}}, we exploit 2 two-layer convolutional blocks as both encoder and decoder to form a simplified Unet-like~\cite{Ronneberger2015U} architecture. However, limited by the locality assumptions, there is a significant deterioration in precision. Similarly, we reconstruct a simplified transformer-based generative module referring to {\cite{seg2021}}, whose encoder and decoder both consist of two transformer layers.} In this case, the performance across all metrics is worse than ours, but the complexity exceeds. These results demonstrate our generative module is suitable for plunging into a transformer to generate simulated cross-view knowledge.

\begin{table}[htb]
\caption{\textbf{Detailed ablation study} of the composition of Generative Module. }
    \begin{adjustwidth}{-\extralength}{0cm}
    \newcolumntype{C}{>{\centering\arraybackslash}X}
    \newcolumntype{M}[1]{>{\centering\arraybackslash}m{#1}}
    \begin{tabularx}{\fulllength}{M{2cm}CCCCCCCCCC}
        \toprule
        \multirow{2}{*}{\tabincell{c}{\textbf{Method}}}&
        \multicolumn{2}{c}{\tabincell{c}{\textbf{Complexity}}} & 
        \multicolumn{4}{c}{\tabincell{c}{\textbf{CVUSA}}} &
        \multicolumn{4}{c}{\tabincell{c}{\textbf{CVACT\_val}}}\\
        & GFLOPs$\downarrow$& Param.$\downarrow$ &{r@1} & {r@5} & {r@10} & {r@1\%} & {r@1} & {r@5} & {r@10} & {r@1\%} \\
        \midrule
        VAE~\cite{kingma2013auto}     &59.89    &141.84M &94.11  &98.27  &99.03  &99.71  &85.32  &94.42  &96.04  &98.41\\
        Unet~\cite{Ronneberger2015U} &59.53 &134.74M &92.04  &97.91  &98.80  &99.67  &82.31  &93.08  &95.09  &98.30\\
        Transformer~\cite{A2017Attention}&61.20 &276.49M &94.37  &98.30  &99.08  &99.74  &85.35  &94.45  &96.03  &98.44\\
        {\bf Ours}&59.71  &171.97M  &\bf94.50  &\bf98.41  &\bf99.20  &\bf99.78&\bf85.42  &\bf94.64  &\bf96.11  &\bf98.51\\
        \bottomrule
    \end{tabularx}%
    \end{adjustwidth}
  \label{tab:5}%
\end{table}%

{\bf Study of Recurrent Learning Step:} {To determine the best recurrent learning step that balances quality and complexity. We report the results trained with different recurrent steps in Table~\ref{tab:6}. 
This demonstrates that increasing the recurrent learning step leads to improved performance, which proves that recurrent learning can fully mine the representation ability of generative cross-view knowledge. As the learning step increases from 3 to 6, top 1(r@1) retrieval accuracy improves significantly. 
However, the gains are negligible and even degrade as the recurrent step rises to 9. We analyze the quantity and quality of the new generative knowledge becomes more difficult as the recurrent step rises. Therefore, The recurrent learning step is set to 6 to achieve a trade-off between accuracy and time cost.}

\begin{table}[htp]
\caption{\textbf{Detailed ablation study} of different parameter settings.}
    \newcolumntype{C}{>{\centering\arraybackslash}X}
    \begin{tabularx}{\textwidth}{CCCCCCCCC}
    \toprule
    \multirow{2}{*}{\tabincell{c}{\textbf{CVI}}}&
    \multicolumn{4}{c}{\tabincell{c}{\textbf{CVUSA}}}&
    \multicolumn{4}{c}{\tabincell{c}{\textbf{CVACT\_val}}}\\
    &  {r@1} & {r@5} & {r@10} & {r@1\%} & {r@1} & {r@5} & {r@10} & {r@1\%}\\
    \midrule
    L = 1     &87.07  &96.48  &97.72  &99.65  &77.92 &90.85 &93.27 &97.20 \\
    L = 3     &89.67  &97.15  &98.26  &99.70  &80.40 &91.63 &94.18 &98.17 \\
    L = 6     &\bf94.50  &\bf98.41  &\bf99.20  &\bf99.78  &\bf85.42  &\bf94.64  &96.11  &\bf98.51  \\
    L = 9     &94.25  &98.39  &99.18  &99.76   &85.20  &94.61 &\bf96.12 &98.49 \\
    \bottomrule
  \end{tabularx}%
  \label{tab:6}%
\end{table}%


\subsection{Supplementary Experiment}

To further explore the rationality of each module, \emph{i.e.}, multi-scale feature aggregation (MSFA), spatial context enhancement (SCE), and spatial attention (SA) in cascaded attention masking (CAMask) algorithm. We conduct intensive experiments with different settings on both \emph{CVUSA}~\cite{workman2015wide} and \emph{CVACT\_val}~\cite{liu2019lending}, and all results are reported in Table~\ref{tab:7}.

{\bf Exploration of MSFA:} We redesign a parallel multi-branch convolutional module named SCE. Unlike existing works, we introduce a novel feature aggregation tactic named MSFA to further integrate branches with different scales. To prove the necessity of MSFA, we replace MSFA with direct addition operations and convolutional layers, respectively. We observe that all the metrics decrease and the top 1(r@1) retrieval accuracy decreases drastically by more than 1\%. {To further illustrate the advancements of our proposed MSFA, we select five typical feature aggregation tactics, including squeeze-and-excitation networks (SENet)~\cite{H2018Squeeze}, convolutional block attention module (Cbam) ~\cite{woo2018cbam}, self-calibrated convolution (SCNet)~\cite{liu2020Improving}, Non\_local ~\cite{wang2018non}, and selective kernel convolution (SKC)~\cite{li2020Selective}, and then plug them into SCE, respectively. As shown in Table~\ref{tab:7} (Above), SCE with MSFA achieves the best retrieval accuracy with a parameter increase of less than 10M. }

{\bf Exploration of SCE:} {To illustrate the superiority of our proposed SCE, we select 2 typical multi-branch convolutional modules, \emph{i.e.} Texture-enhanced module (TEM)~\cite{fan2021concealed} and Receptive field block (RFB)}~\cite{liu2018receptive}. TEM~\cite{fan2021concealed} aims to capture fine-grained texture and context features, which was initially employed in the concealed object detection (COD) task. Inspired by the human visual system, RFB~\cite{liu2018receptive} introduced a multi-branch dilated convolution to enhance the feature extraction capability of the network. Building on this purpose, we introduce SCE with MSFA. As shown in Table~\ref{tab:7} (Middle), SCE achieves the optimal performance with fewer parameters, suggesting the SCE equipped with an attention-based feature aggregation tactic is more suitable for the CVGL task. 

\begin{table}[t!]
\caption{{\bf Detailed ablation study} of the rationality of MSFA and MSCM. }
    \begin{adjustwidth}{-\extralength}{0cm}
    \newcolumntype{C}{>{\centering\arraybackslash}X}
    \newcolumntype{M}[1]{>{\centering\arraybackslash}m{#1}}
    \begin{tabularx}{\fulllength}{M{2cm}CCCCCCCCCC} 
        \toprule
        \multirow{2}{*}{\textbf{Module}}&
        \multicolumn{2}{c}{\tabincell{c}{\textbf{Complexity}}} & 
        \multicolumn{4}{c}{\tabincell{c}{\textbf{CVUSA}}} &
        \multicolumn{4}{c}{\tabincell{c}{\textbf{CVACT\_val}}}\\
        & GFLOPs$\downarrow$   & Param.$\downarrow$ & {r@1} & {r@5} & {r@10} & {r@1\%} & {r@1} & {r@5} & {r@10} & {r@1\%} \\
        \midrule
        \multicolumn{10}{l}{{Why MSFA in SCE? - Comparison with other aggregation methods.}} \\
        \midrule
        {add}    & 54.62   &161.30M &93.11  &98.22  &98.93   &99.75  &83.55  &93.66  &95.56 &98.26\\
        {concat+conv}&  55.75  &163.71M &92.62  &97.96  &98.85   &99.72  &83.07  &93.84  &95.57 &98.21 \\
        {SENet~\cite{H2018Squeeze}}  & 54.63   &161.95M &93.48  &98.31  &98.99   &99.74  &85.01  &94.40  &96.02 &98.48\\
        {Cbam~\cite{woo2018cbam}}    &  54.63 &161.96M &93.61  &98.39  &99.02   &99.73  &84.89  &94.27  &96.03 &98.43\\
        {SCNet~\cite{liu2020Improving}}   & 56.68  &166.12M &93.82  &98.38  &99.08   &99.74  &84.67  &94.28  &95.97 &98.35\\
        {Non\_Local~\cite{wang2018non}}&  65.36  &163.72M &92.84  &98.01  &98.86   &99.67  &83.48  &93.99  &95.68 &98.29\\
        {SKC~\cite{li2020Selective}}    &  73.84  &201.91M &93.92  &98.36  &99.03   &99.78  &84.95  &94.36  &95.81 &98.44\\
        \midrule
        {\bf MSFA} & 59.71 &171.97M &\bf94.50  &\bf98.41  &\bf99.20  &\bf99.78  &\bf85.42  &\bf94.64  &\bf96.03  &\bf98.51\\
        \midrule
        \multicolumn{8}{l}{{Why SCE in CAMask? - Comparison with other parallel multi-branch convolutional modules.}} \\
        \midrule
        {Conv}    &  76.38 &208.51M &92.97 &98.16 &98.87 &99.72  &83.51  &94.02 &95.78 &98.37 \\
        {TEM~\cite{fan2021concealed}}    & 67.05   &187.31M &94.21 &98.37 &99.05 &99.74  &85.10  &94.56 &95.99 &98.43\\
        {RFB~\cite{liu2018receptive}}    & 67.61   &188.47M &94.13 &98.33 &99.07 &99.72  &85.08  &94.48 &96.01 &98.40\\
        \midrule
        {\bf SCE} & 59.71 &171.97M &\bf94.50  &\bf98.41  &\bf99.20  &\bf99.78  &\bf85.42  &\bf94.64  &\bf96.03  &\bf98.51\\
        \midrule
        \multicolumn{10}{l}{{Why SA in CAMask? - Comparison with other attention masks generation methods}} \\
        \midrule
        {GAP}   &   59.71  &171.97M   &93.62  &98.41  &99.07  &99.73  &84.33  &94.31  &95.67 &98.43\\
        {GMP}     &  59.71 &171.97M   &93.43  &98.33  &99.05  &99.75  &84.20  &94.21  &95.88  &98.45\\

        \midrule
        {\bf SA} & 59.71 &171.97M &\bf94.50  &\bf98.41  &\bf99.20  &\bf99.78  &\bf85.42  &\bf94.64  &\bf96.03  &\bf98.51\\
        \bottomrule
    \end{tabularx}%
    \end{adjustwidth}
  \label{tab:7}%
\end{table}%

{\bf Exploration of SA:} {Spatial attention has been proposed to adaptively learn discriminative regions in the feature map to generate the spatial masks. We use SA to connect cascaded structures and use spatial masks generated from cross-level semantic information to compensate for the loss of spatial information due to reduced spatial resolutions via multiplying with high-level semantic features.} To study the effectiveness of SA, we replace SA with global average/max pooling (GAP/GMP) layers and observe an overall decrease across all metrics in which top 1(r@1) retrieval accuracy suffers drastic decreases by 0.88\% (94.50\% $\rightarrow$ 93.62\%) and 1.07\% (94.50\% $\rightarrow$ 93.43\%), showcasing the necessity and effectiveness of the SA mechanism in CAMask.






\section{Discussion}\label{sec:5}


As described in Sec.~\ref{sec:4}, mutual generative transformer learning (MGTL) outperforms recent outstanding cross-view geo-localization (CVGL) works significantly across almost all of the metrics on widely-used benchmarks \emph{CVUSA}~\cite{workman2015wide} and \emph{CVACT}~\cite{liu2019lending}, owing to our cascaded attention masking (CAMask) algorithm and cross-view interaction (CVI) tactic. CAMask is integrated into feature extractor VGG16~\cite{simonyan2014very} to encourage co-visual regions for reasoning  during generative transformer learning, which eradicates the interference of viewpoint-sensitive regions. CVI is implemented by cross-view generative module and generative knowledge supported transformer learning. Cross-view mutual generative learning aims to simulate the feature representations across views. Subsequently, exploiting the generative knowledge to mine the semantic consistency through the attention mechanism in recurrent transformer learning. Our findings perform excellently in cross-view image matching essential for CVGL. In addition, our MGTL enhances the generalizability of CVGL, driving vision-based geo-localization solutions applicated in autonomous driving fields without GPS support.

\section{Future Work}\label{sec:7}
{By exploiting the inter-view semantic consistency, mutual learning can alleviate ambiguity in cross-view matching. The study of view matching in UAV localization has also been conducted in a similar area for the purpose of completing UAV geographic localization. A benchmark called University-1652~\cite{zheng2020university} aims to establish correspondence between a UAV view and a satellite view. In our further work, we will explore how mutual learning techniques can play a role in this similar field, including: \textbf{1)} The slight difference in view perspective between the satellite view and UAV view makes the cross-view semantic consistency easier to obtain, allowing mutual learning to go further in enhancing semantics. \textbf{2)} Shared parameter learning, which can make the network more efficient, should be explored in the context of mutual learning.}


\section{Conclusions}\label{sec:6}

This paper proposes a novel mutual generative transformer learning network, denoted as MGTL, for addressing the cross-view geo-localization problem.
Existing methods commonly rely on a CNN-based Siamese-like backbone to extract high-order feature representations and treat each region equally. Viewpoint-sensitive regions with drastic appearance differences, however, hinder image matching significantly. Using a cascaded attention masking algorithm, we introduce a spatial context enhancement module and a spatial attention module in the VGG16 to capture co-visual information. As for semantic consistency learning, it is hardly examined in recent works, but incorporating consistency constraints by cross-view interaction during the recurrent learning process will benefit similarity computing. To facilitate high-order information mining within each view, we construct cross-view generative modules and inject their generative cross-view knowledge into a transformer-based framework. Extensive qualitative and quantitative experiments demonstrate that mutual generative transformer learning  significantly alleviates the impact of spatial information mismatch caused by drastic viewpoint changes. By examining cross-view interaction, we highlight the potential of this perspective to advance automobile geolocation identification research in GPS-denied conditions. 

\authorcontributions{Conceptualization, Jianwei Zhao and Qiang Zhai; Data curation, Jianwei Zhao and Qiang Zhai; Formal analysis, Pengbo Zhao and Rui Huang; Funding acquisition, Qiang Zhai and Hong Cheng; Investigation, Qiang Zhai and Rui Huang; Methodology, Qiang Zhai; Project administration, Qiang Zhai and Hong Cheng; Resources, Qiang Zhai; Software, Jianwei Zhao; Supervision, Qiang Zhai and Hong Cheng; Validation, Jianwei Zhao and Qiang Zhai; Visualization, Rui Huang; Writing – original draft, Jianwei Zhao; Writing – review \& editing, Qiang Zhai, Pengbo Zhao and Rui Huang. All authors have read and agreed to the published version of the manuscript.}

\funding{\added[id=A.,comment={Funding updating.}]{This research was funded by the National Key Research and Development Program of China (NO. 2022YFB2503004) and the National Natural Science Foundation of China (NSFC) (NO. U1964203).}}

\acknowledgments{The authors would like to thank all experts in the robotics and remote sensing community for their contribution to cross-view geo-localization.}

\conflictsofinterest{The authors declare no conflict of interest.}














\begin{adjustwidth}{-\extralength}{0cm}

\reftitle{References}


\bibliography{main}

\begin{thebibliography}{999}

\bibitem[Saurer \em{et~al.}(2016)Saurer, Baatz, K{\"o}ser, Pollefeys,
  et~al.]{saurer2016image}
Saurer, O.; Baatz, G.; K{\"o}ser, K.; Pollefeys, M.;  et~al.
\newblock Image based geo-localization in the alps.
\newblock {\em International Journal of Computer Vision} {\bf 2016}, {\em
  116},~213--225.

\bibitem[Senlet and Elgammal(2012)]{2012Satellite}
Senlet, T.; Elgammal, A.
\newblock Satellite image-based precise robot localization on sidewalks.
\newblock In Proceedings of the IEEE International Conference on Robotics and
  Automation,  2012, pp. 2647--2653.

\bibitem[Xiao \em{et~al.}(2020)Xiao, Codevilla, Gurram, Urfalioglu, and
  L{\'o}pez]{xiao2020multimodal}
Xiao, Y.; Codevilla, F.; Gurram, A.; Urfalioglu, O.; L{\'o}pez, A.M.
\newblock Multimodal end-to-end autonomous driving.
\newblock {\em IEEE Transactions on Intelligent Transportation Systems} {\bf
  2020}, {\em 23},~537--547.
\newblock \mbox{doi}: \url{10.1109/TITS.2020.3013234}.

\bibitem[Wang \em{et~al.}(2022)Wang, Zhang, and Li]{wang2022satellite}
Wang, S.; Zhang, Y.; Li, H.
\newblock Satellite image based cross-view localization for autonomous vehicle.
\newblock {\em arXiv preprint arXiv:2207.13506} {\bf 2022}.

\bibitem[Thoma \em{et~al.}(2019)Thoma, Paudel, Chhatkuli, Probst, and
  Gool]{thoma2019mapping}
Thoma, J.; Paudel, D.P.; Chhatkuli, A.; Probst, T.; Gool, L.V.
\newblock Mapping, localization and path planning for image-based navigation
  using visual features and map.
\newblock In Proceedings of the IEEE/CVF International Conference on Computer
  Vision,  2019, pp. 7383--7391.

\bibitem[Roy and Debarshi(2020)]{roy2020uav}
Roy, N.; Debarshi, S.
\newblock Uav-based person re-identification and dynamic image routing using
  wireless mesh networking.
\newblock In Proceedings of the 2020 7th International Conference on Signal
  Processing and Integrated Networks (SPIN). IEEE,  2020, pp. 914--917.

\bibitem[Hu and Lee(2020)]{hu2020image}
Hu, S.; Lee, G.H.
\newblock Image-based geo-localization using satellite imagery.
\newblock {\em IJCV} {\bf 2020}, {\em 128},~1205--1219.

\bibitem[Arandjelovic \em{et~al.}(2016)Arandjelovic, Gronat, Torii, Pajdla, and
  Sivic]{arandjelovic2016netvlad}
Arandjelovic, R.; Gronat, P.; Torii, A.; Pajdla, T.; Sivic, J.
\newblock NetVLAD: CNN architecture for weakly supervised place recognition.
\newblock In Proceedings of the IEEE/CVF Conference on Computer Vision and
  Pattern Recognition,  2016, pp. 5297--5307.

\bibitem[Workman and Jacobs(2015)]{workman2015location}
Workman, S.; Jacobs, N.
\newblock On the location dependence of convolutional neural network features.
\newblock In Proceedings of the IEEE/CVF Winter Conference on Computer Vision
  and Pattern Recognition,  2015, pp. 70--78.

\bibitem[Vo and Hays(2016)]{vo2016localizing}
Vo, N.N.; Hays, J.
\newblock Localizing and orienting street views using overhead imagery.
\newblock In Proceedings of the European Conference on Computer Vision.
  Springer,  2016, pp. 494--509.

\bibitem[Hu \em{et~al.}(2018)Hu, Feng, Nguyen, and Lee]{hu2018cvm}
Hu, S.; Feng, M.; Nguyen, R.M.; Lee, G.H.
\newblock Cvm-net: Cross-view matching network for image-based ground-to-aerial
  geo-localization.
\newblock In Proceedings of the IEEE/CVF Conference on Computer Vision and
  Pattern Recognition,  2018, pp. 7258--7267.

\bibitem[Regmi and Shah(2019)]{regmi2019bridging}
Regmi, K.; Shah, M.
\newblock Bridging the domain gap for ground-to-aerial image matching.
\newblock In Proceedings of the IEEE/CVF Conference on Computer Vision and
  Pattern Recognition,  2019, pp. 470--479.

\bibitem[Zhu \em{et~al.}(2022)Zhu, Shah, and Chen]{zhu2022transgeo}
Zhu, S.; Shah, M.; Chen, C.
\newblock TransGeo: Transformer Is All You Need for Cross-view Image
  Geo-localization.
\newblock In Proceedings of the IEEE/CVF Conference on Computer Vision and
  Pattern Recognition,  2022, pp. 1162--1171.

\bibitem[Yang \em{et~al.}(2021)Yang, Lu, and Zhu]{yang2021cross}
Yang, H.; Lu, X.; Zhu, Y.
\newblock Cross-view Geo-localization with Layer-to-Layer Transformer.
\newblock {\em Advances in Neural Information Processing Systems} {\bf 2021},
  {\em 34},~29009--29020.

\bibitem[Chen \em{et~al.}(2014)Chen, Lam, Jacobson, and
  Milford]{chen2014convolutional}
Chen, Z.; Lam, O.; Jacobson, A.; Milford, M.
\newblock Convolutional neural network-based place recognition.
\newblock  2014.

\bibitem[Xin \em{et~al.}(2019)Xin, Cai, Lu, Xing, Cai, Zhang, Yang, and
  Wang]{Xin2019LLRN}
Xin, Z.; Cai, Y.; Lu, T.; Xing, X.; Cai, S.; Zhang, J.; Yang, Y.; Wang, Y.
\newblock Localizing Discriminative Visual Landmarks for Place Recognition.
\newblock In Proceedings of the IEEE International Conference on Robotics and
  Automation,  2019, pp. 5979--5985.

\bibitem[Khaliq \em{et~al.}()Khaliq, Milford, and Garg]{Khaliq2022MultiRes}
Khaliq, A.; Milford, M.; Garg, S.
\newblock MultiRes-NetVLAD: Augmenting Place Recognition Training With
  Low-Resolution Imagery.
\newblock {\em IEEE Robotics and Automation Letters}, pp. 3882--3889.

\bibitem[Yu \em{et~al.}(2019)Yu, Zhu, Zhang, Huang, and Tao]{yu2019spatial}
Yu, J.; Zhu, C.; Zhang, J.; Huang, Q.; Tao, D.
\newblock Spatial pyramid-enhanced NetVLAD with weighted triplet loss for place
  recognition.
\newblock {\em IEEE Transactions on neural networks and learning systems} {\bf
  2019}, {\em 31},~661--674.
\newblock \mbox{doi}: \url{10.1109/TNNLS.2019.2908982}.

\bibitem[Latif \em{et~al.}(2018)Latif, Garg, Milford, and
  Reid]{latif2018addressing}
Latif, Y.; Garg, R.; Milford, M.; Reid, I.
\newblock Addressing challenging place recognition tasks using generative
  adversarial networks.
\newblock In Proceedings of the IEEE European Conference on Computer Vision.
  IEEE,  2018, pp. 2349--2355.

\bibitem[Castaldo \em{et~al.}(2015)Castaldo, Zamir, Angst, Palmieri, and
  Savarese]{2015Semantic}
Castaldo, F.; Zamir, A.; Angst, R.; Palmieri, F.; Savarese, S.
\newblock Semantic cross-view matching.
\newblock In Proceedings of the IEEE/CVF International Conference on Computer
  Vision Workshops,  2015, pp. 9--17.

\bibitem[Mousavian and Kosecka(2016)]{2016Semantic}
Mousavian, A.; Kosecka, J.
\newblock Semantic Image Based Geolocation Given a Map,  2016.

\bibitem[Zhu \em{et~al.}(2021)Zhu, Yang, and Chen]{zhu2021vigor}
Zhu, S.; Yang, T.; Chen, C.
\newblock Vigor: Cross-view image geo-localization beyond one-to-one retrieval.
\newblock In Proceedings of the IEEE/CVF Conference on Computer Vision and
  Pattern Recognition,  2021, pp. 3640--3649.

\bibitem[Shi \em{et~al.}(2019)Shi, Liu, Yu, and Li]{shi2019spatial}
Shi, Y.; Liu, L.; Yu, X.; Li, H.
\newblock Spatial-aware feature aggregation for image based cross-view
  geo-localization.
\newblock {\em Advances in Neural Information Processing Systems} {\bf 2019},
  {\em 32}.

\bibitem[Shi \em{et~al.}(2020)Shi, Yu, Liu, Zhang, and Li]{shi2020optimal}
Shi, Y.; Yu, X.; Liu, L.; Zhang, T.; Li, H.
\newblock Optimal feature transport for cross-view image geo-localization {\bf
  2020}.
\newblock {\em 34},~11990--11997.

\bibitem[Wang \em{et~al.}(2022)Wang, Fan, Liu, and Sun]{wang2022transgcnn}
Wang, T.; Fan, S.; Liu, D.; Sun, C.
\newblock Transformer-Guided Convolutional Neural Network for Cross-View
  Geolocalization.
\newblock {\em arXiv preprint arXiv:2204.09967} {\bf 2022}.

\bibitem[Wang \em{et~al.}(2021)Wang, Zheng, Yan, Zhang, Sun, Zheng, and
  Yang]{2021Each}
Wang, T.; Zheng, Z.; Yan, C.; Zhang, J.; Sun, Y.; Zheng, B.; Yang, Y.
\newblock Each part matters: Local patterns facilitate cross-view
  geo-localization.
\newblock {\em IEEE Transactions on Circuits and Systems for Video Technology}
  {\bf 2021}, {\em 32},~867--879.
\newblock \mbox{doi}: \url{10.1109/TCSVT.2021.3061265}.

\bibitem[Wang \em{et~al.}(2022)Wang, Zheng, Zhu, Gao, Yang, and
  Yan]{wang2022learning}
Wang, T.; Zheng, Z.; Zhu, Z.; Gao, Y.; Yang, Y.; Yan, C.
\newblock Learning Cross-view Geo-localization Embeddings via Dynamic Weighted
  Decorrelation Regularization.
\newblock {\em arXiv preprint arXiv:2211.05296} {\bf 2022}.

\bibitem[Zhu \em{et~al.}(2023)Zhu, Yang, Lu, and Huang]{zhu2023simple}
Zhu, Y.; Yang, H.; Lu, Y.; Huang, Q.
\newblock Simple, Effective and General: A New Backbone for Cross-view Image
  Geo-localization.
\newblock {\em arXiv preprint arXiv:2302.01572} {\bf 2023}.

\bibitem[Zhang \em{et~al.}(2022)Zhang, Li, Sultani, Zhou, and
  Wshah]{zhang2022cross}
Zhang, X.; Li, X.; Sultani, W.; Zhou, Y.; Wshah, S.
\newblock Cross-view Geo-localization via Learning Disentangled Geometric
  Layout Correspondence.
\newblock {\em arXiv preprint arXiv:2212.04074} {\bf 2022}.

\bibitem[Workman \em{et~al.}(2015)Workman, Souvenir, and
  Jacobs]{workman2015wide}
Workman, S.; Souvenir, R.; Jacobs, N.
\newblock Wide-area image geolocalization with aerial reference imagery.
\newblock In Proceedings of the IEEE/CVF Conference on Computer Vision and
  Pattern Recognition,  2015, pp. 3961--3969.

\bibitem[Liu and Li(2019)]{liu2019lending}
Liu, L.; Li, H.
\newblock Lending orientation to neural networks for cross-view
  geo-localization.
\newblock In Proceedings of the IEEE/CVF Conference on Computer Vision and
  Pattern Recognition,  2019, pp. 5624--5633.

\bibitem[Zhu \em{et~al.}(2021)Zhu, Sun, Lu, and Jia]{zhu2021geographic}
Zhu, Y.; Sun, B.; Lu, X.; Jia, S.
\newblock Geographic Semantic Network for Cross-View Image Geo-Localization.
\newblock {\em IEEE Transactions on Geoscience and Remote Sensing} {\bf 2021},
  {\em 60},~1--15.
\newblock \mbox{doi}: \url{10.1109/TGRS.2021.3121337}.

\bibitem[Zhu \em{et~al.}(2023)Zhu, Yang, Dai, Fan, and Ye]{zhu2022r2fd2}
Zhu, B.; Yang, C.; Dai, J.; Fan, J.; Ye, Y.
\newblock R2FD2: Fast and Robust Matching of Multimodal Remote Sensing Image
  via Repeatable Feature Detector and Rotation-invariant Feature Descriptor.
\newblock {\em IEEE Transactions on Geoscience and Remote Sensing} {\bf 2023}.
\newblock \mbox{doi}: \url{10.1109/TGRS.2023.3264610}.

\bibitem[Regmi and Borji(2018)]{regmi2018cross}
Regmi, K.; Borji, A.
\newblock Cross-view image synthesis using conditional gans.
\newblock In Proceedings of the IEEE/CVF Conference on Computer Vision and
  Pattern Recognition,  2018, pp. 3501--3510.

\bibitem[Lu \em{et~al.}(2020)Lu, Li, Cui, Oswald, Pollefeys, and
  Qin]{lu2020geometry}
Lu, X.; Li, Z.; Cui, Z.; Oswald, M.R.; Pollefeys, M.; Qin, R.
\newblock Geometry-aware satellite-to-ground image synthesis for urban areas.
\newblock In Proceedings of the IEEE/CVF Conference on Computer Vision and
  Pattern Recognition,  2020, pp. 859--867.

\bibitem[Ding \em{et~al.}(2020)Ding, Wu, Tang, Wu, Gao, and
  Jing]{ding2020cross}
Ding, H.; Wu, S.; Tang, H.; Wu, F.; Gao, G.; Jing, X.Y.
\newblock Cross-view image synthesis with deformable convolution and attention
  mechanism.
\newblock In Proceedings of the Chinese Conference on Pattern Recognition and
  Computer Vision (PRCV). Springer,  2020, pp. 386--397.

\bibitem[Lin \em{et~al.}(2015)Lin, Cui, Belongie, and Hays]{Lin2015Learning}
Lin, T.Y.; Cui, Y.; Belongie, S.; Hays, J.
\newblock Learning deep representations for ground-to-aerial geolocalization.
\newblock In Proceedings of the IEEE/CVF Conference on Computer Vision and
  Pattern Recognition,  2015, pp. 5007--5015.

\bibitem[Sun \em{et~al.}(2019)Sun, Chen, Zhu, and Jiang]{sun2019GeoCapsNet}
Sun, B.; Chen, C.; Zhu, Y.; Jiang, J.
\newblock GeoCapsNet: Aerial to Ground view Image Geo-localization using
  Capsule Network.
\newblock {\em arXiv preprint arXiv:1904.06281} {\bf 2019}.

\bibitem[Cai \em{et~al.}(2019)Cai, Guo, Khan, Hu, and Wen]{2019Ground}
Cai, S.; Guo, Y.; Khan, S.; Hu, J.; Wen, G.
\newblock Ground-to-Aerial Image Geo-Localization With a Hard Exemplar
  Reweighting Triplet Loss.
\newblock In Proceedings of the IEEE/CVF International Conference on Computer
  Vision,  2019, pp. 8391--8400.

\bibitem[Ren \em{et~al.}(2021)Ren, Tang, and Sebe]{ren2021cascaded}
Ren, B.; Tang, H.; Sebe, N.
\newblock Cascaded cross mlp-mixer gans for cross-view image translation.
\newblock {\em arXiv preprint arXiv:2110.10183} {\bf 2021}.

\bibitem[Toker \em{et~al.}(2021)Toker, Zhou, Maximov, and
  Leal-Taix{\'e}]{toker2021coming}
Toker, A.; Zhou, Q.; Maximov, M.; Leal-Taix{\'e}, L.
\newblock Coming down to earth: Satellite-to-street view synthesis for
  geo-localization.
\newblock In Proceedings of the IEEE/CVF International Conference on Computer
  Vision,  2021, pp. 6488--6497.

\bibitem[Vaswani \em{et~al.}(2017)Vaswani, Shazeer, Parmar, Uszkoreit, Jones,
  Gomez, Kaiser, and Polosukhin]{A2017Attention}
Vaswani, A.; Shazeer, N.; Parmar, N.; Uszkoreit, J.; Jones, L.; Gomez, A.N.;
  Kaiser, L.; Polosukhin, I.
\newblock Attention Is All You Need.
\newblock  2017, Vol.~30.

\bibitem[Dosovitskiy \em{et~al.}(2020)Dosovitskiy, Beyer, Kolesnikov,
  Weissenborn, Zhai, Unterthiner, Dehghani, Minderer, Heigold, Gelly,
  et~al.]{dosovitskiy2020image}
Dosovitskiy, A.; Beyer, L.; Kolesnikov, A.; Weissenborn, D.; Zhai, X.;
  Unterthiner, T.; Dehghani, M.; Minderer, M.; Heigold, G.; Gelly, S.;  et~al.
\newblock An image is worth 16x16 words: Transformers for image recognition at
  scale.
\newblock {\em arXiv preprint arXiv:2010.11929} {\bf 2020}.

\bibitem[Chen \em{et~al.}(2021)Chen, Wang, Guo, Xu, Deng, Liu, Ma, Xu, Xu, and
  Gao]{chen2021pre}
Chen, H.; Wang, Y.; Guo, T.; Xu, C.; Deng, Y.; Liu, Z.; Ma, S.; Xu, C.; Xu, C.;
  Gao, W.
\newblock Pre-trained image processing transformer.
\newblock In Proceedings of the IEEE/CVF Conference on Computer Vision and
  Pattern Recognition,  2021, pp. 12299--12310.

\bibitem[Bhojanapalli \em{et~al.}(2021)Bhojanapalli, Chakrabarti, Glasner, Li,
  Unterthiner, and Veit]{bhojanapalli2021understanding}
Bhojanapalli, S.; Chakrabarti, A.; Glasner, D.; Li, D.; Unterthiner, T.; Veit,
  A.
\newblock Understanding robustness of transformers for image classification.
\newblock In Proceedings of the IEEE/CVF International Conference on Computer
  Vision,  2021, pp. 10231--10241.

\bibitem[Lanchantin \em{et~al.}(2021)Lanchantin, Wang, Ordonez, and
  Qi]{lanchantin2021general}
Lanchantin, J.; Wang, T.; Ordonez, V.; Qi, Y.
\newblock General multi-label image classification with transformers.
\newblock In Proceedings of the IEEE/CVF Conference on Computer Vision and
  Pattern Recognition,  2021, pp. 16478--16488.

\bibitem[Strudel \em{et~al.}(2021)Strudel, Pinel, Laptev, and Schmid]{seg2021}
Strudel, R.; Pinel, R.G.; Laptev, I.; Schmid, C.
\newblock Segmenter: Transformer for Semantic Segmentation.
\newblock In Proceedings of the IEEE/CVF International Conference on Computer
  Vision,  2021, pp. 7262--7272.

\bibitem[Jin \em{et~al.}(2021)Jin, Han, and Ko]{jin2021trseg}
Jin, Y.; Han, D.; Ko, H.
\newblock Trseg: Transformer for semantic segmentation.
\newblock {\em Pattern Recognition Letters} {\bf 2021}, {\em 148},~29--35.

\bibitem[Zheng \em{et~al.}(2021)Zheng, Lu, Zhao, Zhu, Luo, Wang, Fu, Feng,
  Xiang, Torr, et~al.]{zheng2021rethinking}
Zheng, S.; Lu, J.; Zhao, H.; Zhu, X.; Luo, Z.; Wang, Y.; Fu, Y.; Feng, J.;
  Xiang, T.; Torr, P.H.;  et~al.
\newblock Rethinking semantic segmentation from a sequence-to-sequence
  perspective with transformers.
\newblock In Proceedings of the IEEE/CVF conference on computer vision and
  pattern recognition,  2021, pp. 6881--6890.

\bibitem[Carion \em{et~al.}(2020)Carion, Massa, Synnaeve, Usunier, Kirillov,
  and Zagoruyko]{carion2020end}
Carion, N.; Massa, F.; Synnaeve, G.; Usunier, N.; Kirillov, A.; Zagoruyko, S.
\newblock End-to-end object detection with transformers.
\newblock In Proceedings of the European Conference on Computer Vision.
  Springer,  2020, pp. 213--229.

\bibitem[Misra \em{et~al.}(2021)Misra, Girdhar, and Joulin]{misra2021end}
Misra, I.; Girdhar, R.; Joulin, A.
\newblock An end-to-end transformer model for 3d object detection.
\newblock In Proceedings of the IEEE/CVF International Conference on Computer
  Vision,  2021, pp. 2906--2917.

\bibitem[Zhu \em{et~al.}(2020)Zhu, Su, Lu, Li, Wang, and
  Dai]{zhu2020deformable}
Zhu, X.; Su, W.; Lu, L.; Li, B.; Wang, X.; Dai, J.
\newblock Deformable detr: Deformable transformers for end-to-end object
  detection.
\newblock {\em arXiv preprint arXiv:2010.04159} {\bf 2020}.

\bibitem[Liang \em{et~al.}(2022)Liang, Wang, Wang, Yang, and
  Zhou]{liang2022light}
Liang, Z.; Wang, Y.; Wang, L.; Yang, J.; Zhou, S.
\newblock Light field image super-resolution with transformers.
\newblock {\em IEEE Signal Processing Letters} {\bf 2022}, {\em 29},~563--567.

\bibitem[Zamir \em{et~al.}(2022)Zamir, Arora, Khan, Hayat, Khan, and
  Yang]{zamir2022restormer}
Zamir, S.W.; Arora, A.; Khan, S.; Hayat, M.; Khan, F.S.; Yang, M.H.
\newblock Restormer: Efficient transformer for high-resolution image
  restoration.
\newblock In Proceedings of the IEEE/CVF Conference on Computer Vision and
  Pattern Recognition,  2022, pp. 5728--5739.

\bibitem[Li \em{et~al.}(2021)Li, Liu, Drenkow, Ding, Creighton, Taylor, and
  Unberath]{li2021revisiting}
Li, Z.; Liu, X.; Drenkow, N.; Ding, A.; Creighton, F.X.; Taylor, R.H.;
  Unberath, M.
\newblock Revisiting stereo depth estimation from a sequence-to-sequence
  perspective with transformers.
\newblock In Proceedings of the IEEE/CVF International Conference on Computer
  Vision,  2021, pp. 6197--6206.

\bibitem[Ding \em{et~al.}(2022)Ding, Yuan, Zhu, Zhang, Liu, Wang, and
  Liu]{ding2022transmvsnet}
Ding, Y.; Yuan, W.; Zhu, Q.; Zhang, H.; Liu, X.; Wang, Y.; Liu, X.
\newblock Transmvsnet: Global context-aware multi-view stereo network with
  transformers.
\newblock In Proceedings of the IEEE/CVF Conference on Computer Vision and
  Pattern Recognition,  2022, pp. 8585--8594.

\bibitem[He \em{et~al.}(2021)He, Chen, and Lin]{he2021spatial}
He, X.; Chen, Y.; Lin, Z.
\newblock Spatial-spectral transformer for hyperspectral image classification.
\newblock {\em Remote Sensing} {\bf 2021}, {\em 13},~498.

\bibitem[Qing \em{et~al.}(2021)Qing, Liu, Feng, and Gao]{qing2021improved}
Qing, Y.; Liu, W.; Feng, L.; Gao, W.
\newblock Improved transformer net for hyperspectral image classification.
\newblock {\em Remote Sensing} {\bf 2021}, {\em 13},~2216.

\bibitem[Sun \em{et~al.}(2022)Sun, Zhao, Zheng, and Wu]{sun2022spectral}
Sun, L.; Zhao, G.; Zheng, Y.; Wu, Z.
\newblock Spectral--spatial feature tokenization transformer for hyperspectral
  image classification.
\newblock {\em IEEE Transactions on Geoscience and Remote Sensing} {\bf 2022},
  {\em 60},~1--14.
\newblock \mbox{doi}: \url{10.1109/TGRS.2022.3144158}.

\bibitem[Zhou \em{et~al.}(2022)Zhou, Tian, Zhang, Huo, Xie, and
  Li]{zhou2022multispectral}
Zhou, H.; Tian, C.; Zhang, Z.; Huo, Q.; Xie, Y.; Li, Z.
\newblock Multispectral fusion transformer network for RGB-thermal urban scene
  semantic segmentation.
\newblock {\em IEEE Geoscience and Remote Sensing Letters} {\bf 2022}, {\em
  19},~1--5.

\bibitem[Chu \em{et~al.}(2021)Chu, Tian, Zhang, Wang, Wei, Xia, and
  Shen]{chu2021conditional}
Chu, X.; Tian, Z.; Zhang, B.; Wang, X.; Wei, X.; Xia, H.; Shen, C.
\newblock Conditional positional encodings for vision transformers.
\newblock {\em arXiv preprint arXiv:2102.10882} {\bf 2021}.

\bibitem[Li \em{et~al.}(2021)Li, Zhang, Cao, Timofte, and
  Van~Gool]{li2021localvit}
Li, Y.; Zhang, K.; Cao, J.; Timofte, R.; Van~Gool, L.
\newblock Localvit: Bringing locality to vision transformers.
\newblock {\em arXiv preprint arXiv:2104.05707} {\bf 2021}.

\bibitem[Chen \em{et~al.}(2021)Chen, Fan, and Panda]{chen2021crossvit}
Chen, C.F.R.; Fan, Q.; Panda, R.
\newblock Crossvit: Cross-attention multi-scale vision transformer for image
  classification.
\newblock In Proceedings of the IEEE/CVF International Conference on Computer
  Vision,  2021, pp. 357--366.

\bibitem[Liu \em{et~al.}(2021)Liu, Lin, Cao, Hu, Wei, Zhang, Lin, and
  Guo]{liu2021swin}
Liu, Z.; Lin, Y.; Cao, Y.; Hu, H.; Wei, Y.; Zhang, Z.; Lin, S.; Guo, B.
\newblock Swin transformer: Hierarchical vision transformer using shifted
  windows.
\newblock In Proceedings of the IEEE/CVF International Conference on Computer
  Vision,  2021, pp. 10012--10022.

\bibitem[Yang \em{et~al.}(2021)Yang, Zhai, Li, Huang, Luo, Cheng, and
  Fan]{yang2021uncertainty}
Yang, F.; Zhai, Q.; Li, X.; Huang, R.; Luo, A.; Cheng, H.; Fan, D.P.
\newblock Uncertainty-guided transformer reasoning for camouflaged object
  detection.
\newblock In Proceedings of the IEEE/CVF International Conference on Computer
  Vision,  2021, pp. 4146--4155.

\bibitem[Wang \em{et~al.}()Wang, Yao, Chen, Cai, He, and Liu]{W2021CrossFormer}
Wang, W.; Yao, L.; Chen, L.; Cai, D.; He, X.; Liu, W.
\newblock CrossFormer: A Versatile Vision Transformer Based on Cross-scale
  Attention.
\newblock {\em CoRR abs/2108.00154}.

\bibitem[Simonyan and Zisserman(2014)]{simonyan2014very}
Simonyan, K.; Zisserman, A.
\newblock Very deep convolutional networks for large-scale image recognition.
\newblock {\em arXiv preprint arXiv:1409.1556} {\bf 2014}.

\bibitem[Fan \em{et~al.}(2021)Fan, Ji, Cheng, and Shao]{fan2021concealed}
Fan, D.P.; Ji, G.P.; Cheng, M.M.; Shao, L.
\newblock Concealed object detection.
\newblock {\em IEEE Transactions on Pattern Analysis and Machine Intelligence}
  {\bf 2021}, {\em 44},~6024--6042.
\newblock \mbox{doi}: \url{10.1109/TPAMI.2021.3085766}.

\bibitem[Woo \em{et~al.}(2018)Woo, Park, Lee, and Kweon]{woo2018cbam}
Woo, S.; Park, J.; Lee, J.Y.; Kweon, I.S.
\newblock Cbam: Convolutional block attention module.
\newblock In Proceedings of the European Conference on Computer Vision,  2018,
  pp. 3--19.

\bibitem[Ronneberger \em{et~al.}(2015)Ronneberger, Fischer, and
  Brox]{Ronneberger2015U}
Ronneberger, O.; Fischer, P.; Brox, T.
\newblock U-Net: Convolutional Networks for Biomedical Image Segmentation.
\newblock In Proceedings of the Medical Image Computing and Computer-Assisted
  Intervention. Springer,  2015, pp. 234--241.

\bibitem[Zhai \em{et~al.}(2017)Zhai, Bessinger, Workman, and
  Jacobs]{zhai2017predicting}
Zhai, M.; Bessinger, Z.; Workman, S.; Jacobs, N.
\newblock Predicting ground-level scene layout from aerial imagery.
\newblock In Proceedings of the IEEE Conference on Computer Vision and Pattern
  Recognition,  2017, pp. 867--875.

\bibitem[Deng \em{et~al.}(2009)Deng, Dong, Socher, Li, Li, and
  Fei-Fei]{deng2009imagenet}
Deng, J.; Dong, W.; Socher, R.; Li, L.J.; Li, K.; Fei-Fei, L.
\newblock Imagenet: A large-scale hierarchical image database.
\newblock In Proceedings of the IEEE/CVF Conference on Computer Vision and
  Pattern Recognition,  2009, pp. 248--255.

\bibitem[Shi \em{et~al.}(2020)Shi, Yu, Campbell, and Li]{shi2020looking}
Shi, Y.; Yu, X.; Campbell, D.; Li, H.
\newblock Where am i looking at? joint location and orientation estimation by
  cross-view matching.
\newblock In Proceedings of the IEEE/CVF Conference on Computer Vision and
  Pattern Recognition,  2020, pp. 4064--4072.

\bibitem[Lin \em{et~al.}(2022)Lin, Zheng, Zhong, Luo, Li, Yang, and
  Sebe]{Lin2022usam}
Lin, J.; Zheng, Z.; Zhong, Z.; Luo, Z.; Li, S.; Yang, Y.; Sebe, N.
\newblock Joint Representation Learning and Keypoint Detection for Cross-View
  Geo-Localization.
\newblock {\em IEEE Transactions on Image Processing} {\bf 2022}, {\em
  31},~3780--3792.
\newblock \mbox{doi}: \url{10.1109/TIP.2022.3175601}.

\bibitem[Kingma and Welling(2013)]{kingma2013auto}
Kingma, D.P.; Welling, M.
\newblock Auto-encoding variational bayes.
\newblock {\em arXiv preprint arXiv:1312.6114} {\bf 2013}.

\bibitem[Jie \em{et~al.}(2018)Jie, Li, and Gang]{H2018Squeeze}
Jie, H.; Li, S.; Gang, S.
\newblock Squeeze-and-Excitation Networks.
\newblock In Proceedings of the IEEE/CVF Conference on Computer Vision and
  Pattern Recognition,  2018, pp. 7132--7141.

\bibitem[Liu \em{et~al.}(2020)Liu, Hou, Cheng, Wang, and
  Feng]{liu2020Improving}
Liu, J.J.; Hou, Q.; Cheng, M.M.; Wang, C.; Feng, J.
\newblock Improving Convolutional Networks With Self-Calibrated Convolutions.
\newblock In Proceedings of the IEEE/CVF Conference on Computer Vision and
  Pattern Recognition,  2020, pp. 10096--10105.

\bibitem[Wang \em{et~al.}(2018)Wang, Girshick, Gupta, and He]{wang2018non}
Wang, X.; Girshick, R.; Gupta, A.; He, K.
\newblock Non-local neural networks.
\newblock In Proceedings of the IEEE/CVF Conference on Computer Vision and
  Pattern Recognition,  2018, pp. 7794--7803.

\bibitem[Li \em{et~al.}(2019)Li, Wang, Hu, and Yang]{li2020Selective}
Li, X.; Wang, W.; Hu, X.; Yang, J.
\newblock Selective Kernel Networks.
\newblock In Proceedings of the IEEE/CVF Conference on Computer Vision and
  Pattern Recognition,  2019, pp. 510--519.

\bibitem[Liu \em{et~al.}(2018)Liu, Huang, et~al.]{liu2018receptive}
Liu, S.; Huang, D.;  et~al.
\newblock Receptive field block net for accurate and fast object detection.
\newblock In Proceedings of the European Conference on Computer Vision,  2018,
  pp. 385--400.

\bibitem[Zheng \em{et~al.}(2020)Zheng, Wei, and Yang]{zheng2020university}
Zheng, Z.; Wei, Y.; Yang, Y.
\newblock University-1652: A multi-view multi-source benchmark for drone-based
  geo-localization.
\newblock In Proceedings of the 28th ACM international conference on
  Multimedia,  2020, pp. 1395--1403.

\end{thebibliography}

%


\end{adjustwidth}
\end{document}